\pdfoutput=1
\documentclass[runningheads]{llncs}

\usepackage{accv}

\usepackage{accvabbrv}
\usepackage{graphicx}
\usepackage{booktabs}
\usepackage{algorithm}
\usepackage{algorithmic}
\usepackage{tcolorbox}
\usepackage{multirow}
\usepackage{enumitem}

\usepackage[accsupp]{axessibility}  

\usepackage[a4paper,width=122mm,height=193mm,top=52mm,left=44mm]{geometry}

\usepackage{hyperref}
\usepackage{orcidlink}

\newcommand{\citeS}{\cite}
\newcommand{\bibliographystyleS}[1]{}
\newcommand{\bibliographyS}[1]{}

\begin{document}

\title{Dominant vs. Dominated: Concept-Level Generative Collapse in Diffusion Models}

\titlerunning{Dominant vs. Dominated}

\author{Hayeon Jeong \and Jong-Seok Lee\thanks{Corresponding author.}}

\authorrunning{H.~Jeong, J.-S.~Lee}

\institute{Yonsei University, Korea\\
\email{\{hayeon.jeong, jong-seok.lee\}@yonsei.ac.kr}}

\maketitle

\begin{abstract}
    Text-to-image diffusion models have attracted significant attention for their ability to generate diverse, high-fidelity images. However, in multi-concept generation, one concept token often dominates the output while others are suppressed—a phenomenon we term the Dominant-vs-Dominated (DvD) imbalance. To systematically study this failure mode, we introduce DominanceBench and examine its underlying causes from both data and internal-mechanistic perspectives.
    Our controlled fine-tuning study, which mimics concept learning during diffusion-model training, shows that concepts learned from visually homogeneous (low-variation) concept-specific training images exhibit stronger dominance when composed with others.
    Cross-attention analysis indicates that dominant tokens concentrate attention in early denoising steps, followed by reduced representation of competing concepts.
    Head-ablation analysis further shows that this dominance is distributed across attention heads rather than localized.
    Overall, these findings characterize DvD as a systematic concept-level failure mode and provide a basis for more reliable and controllable multi-concept generation.
    DominanceBench will be released upon publication.
\end{abstract}

\section{Introduction}
\label{sec:intro}

Text-to-image diffusion models \cite{rombach2022highresolution, podell2024sdxl, esser2024scaling, ramesh2021zeroshot, nichol2022glide, zhang2023adding, jeong2025latent} have achieved remarkable success in generating high-quality images from textual descriptions.
However, ensuring the model's representational fidelity to textual concepts \cite{zhang2023text} remains a fundamental challenge.
Recent research has extensively explored this limitation from complementary perspectives.
One line of work investigates \textit{memorization} \cite{carlini2023extracting, wen2024detectingexplainingmitigatingmemorization, somepalli2023diffusion, ren2024unveiling, somepalli2023understanding, kim2025how, hintersdorf2024finding, chen2024exploring, ross2025geometric, jeon2025understanding}, where models reproduce near-identical images across different random seeds, primarily due to excessive duplication of specific image-prompt pairs in training data.
Another line studies compositional text-to-image generation, aiming to reduce concept interference and binding errors under multi-concept prompts \cite{hertz2023prompttoprompt, ding2024freecustom, brooks2023instructpix2pix, patashnik2025nested}.

\begin{figure}[t]
    \centering

    \includegraphics[width=\linewidth]{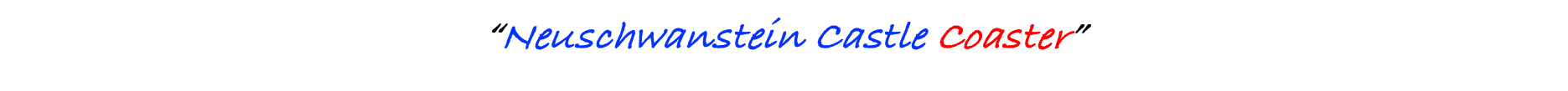}
    \vspace{-1.5em}

    \begin{minipage}{0.49\linewidth}
        \centering
        \includegraphics[width=\linewidth]{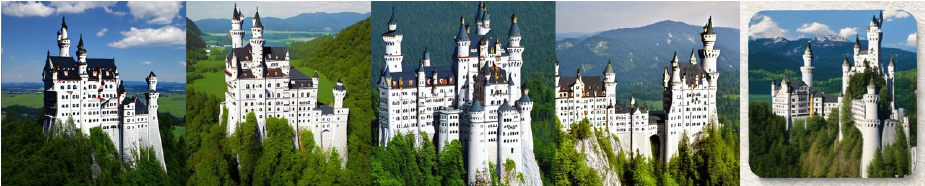}
        \small (a) SD 1.4
    \end{minipage}
    \hfill
    \begin{minipage}{0.49\linewidth}
        \centering
        \includegraphics[width=\linewidth]{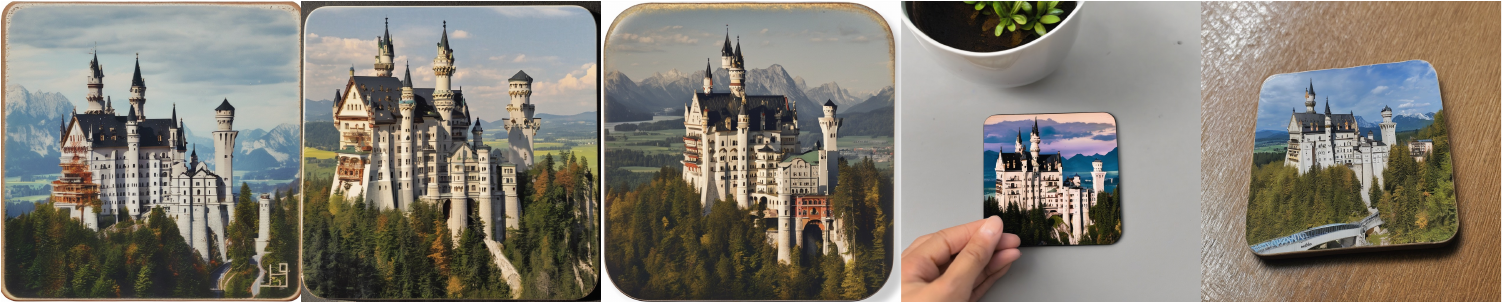}
        \small (b) SDXL
    \end{minipage}

    \vspace{-0.5em}
    \caption{Five-seed generations for ``Neuschwanstein Castle coaster'' using SD 1.4 and SDXL, showing DvD behavior.}
    \vspace{-1.5em}
    \label{fig:dvd_example}
\end{figure}

In this work, we examine a complementary aspect that arises from the \textit{interplay} of these two dimensions—training data characteristics and multi-concept compositional capability.
Specifically, we focus on visual dominance in multi-concept generation: one concept visually dominates the generation while the other is suppressed and rarely appears.
For example, as shown in \cref{fig:dvd_example}, we generate images with Stable Diffusion (SD) v1.4 and SDXL using the prompt ``Neuschwanstein Castle coaster'' with different random seeds.
The castle concept is consistently presented in the generated images, whereas the coaster concept rarely appears.
We refer to this as the \textit{Dominant-vs-Dominated (DvD) phenomenon}.

Motivated by these observations, we study DvD as a concept-level imbalance induced by visual dominance between concepts.
Unlike memorization, which concerns prompt-specific reproduction, and prior work that aims to improve compositional text-to-image generation under multi-concept prompts, we hypothesize that limited variation in a concept's training images (visual homogeneity) can suppress other concepts during multi-concept generation.
We conjecture that it may do so by inducing relatively rigid visual priors that dominate compositions with more diverse concepts.

To test our hypothesis, we conduct controlled fine-tuning studies that simulate concept learning: we learn new concepts from curated training image sets while systematically varying within-concept visual diversity, and observe that dominance increases monotonically as diversity decreases.
To reveal the internal mechanisms of DvD, we analyze cross-attention dynamics and head ablations, revealing that \textit{(1)} DvD prompts exhibit significantly higher attention concentration on dominant tokens in lower-spatial-resolution layers in the U-Net (UNet) at early denoising steps, \textit{(2)} dominated concepts experience rapid attention decay in the early phase of the denoising process, and \textit{(3)} unlike memorization which is localized to specific heads, DvD emerges from distributed cooperation among multiple attention heads.
To support this analysis, we curate DominanceBench (\cref{subsec:db}), a benchmark of 300 two-concept prompts with strong DvD behavior.

Our main contributions are:
\vspace{-0.7em}
\begin{itemize}
    \item We characterize the DvD phenomenon and curate DominanceBench for its systematic evaluation.
    \item We demonstrate through controlled fine-tuning studies on newly learned concepts that limited variation in concept-specific training images plays a causal role in DvD.
    \item We reveal the internal mechanisms of DvD through cross-attention analysis and head ablations, identifying where (lower-spatial-resolution layers), when (early timesteps), and how (distributed across heads) dominance manifests during generation.
\end{itemize}


\section{Related Work}
\label{sec:related_work}
\vspace{-0.5em}
Text-to-image diffusion models have rapidly advanced in fidelity and controllability \cite{ramesh2021zeroshot,nichol2022glide,rombach2022highresolution,podell2024sdxl,esser2024scaling,jeong2025latent,zhang2023adding}, with surveys summarizing this progress \cite{zhang2023text,cao2024controllable,yang2025text2image}.
Two research directions are especially relevant: memorization in diffusion models, and compositional text-to-image generation under multi-concept prompts.
Our work connects these directions by studying concept-level dominance in multi-concept generation and its relationship to training data properties.

\subsection{Memorization in Diffusion Models}
\vspace{-0.5em}
Memorization refers to the reproduction of near-identical training images.
It has been studied from both empirical and mechanistic perspectives \cite{carlini2023extracting,somepalli2023diffusion,somepalli2023understanding}.
Prior work analyzes memorization signals in cross-attention patterns and token-level attention \cite{ren2024unveiling,chen2024exploring}, in prediction magnitudes and detection criteria \cite{wen2024detectingexplainingmitigatingmemorization}, and in localized internal units such as neurons \cite{hintersdorf2024finding}.
Complementary theoretical lenses connect memorization to geometry and sharpness of learned distributions \cite{ross2025geometric,kim2025how,jeon2025understanding}.
We use memorization as a reference point when interpreting DvD mechanisms.

\subsection{Multi-Concept Generation and Evaluation}
\vspace{-0.5em}
Multi-concept prompts pose persistent challenges for text-to-image diffusion models, including concept interference, binding errors, and missing subjects.
A broad line of work seeks to improve compositional generation and editing via attention control and guidance---for example, through attention-based semantic steering \cite{chefer2023attendandexcite}, cross-attention control for edits \cite{hertz2023prompttoprompt,couairon2023diffedit,brooks2023instructpix2pix}, attention-map alignment and control \cite{rassin2023linguistic,wang2024compositional}, and test-time attention segregation \cite{agarwal2023star}.
Other approaches address compositionality through model- and objective-level designs, including composable diffusion and energy-based variants \cite{liu2022compositional,tang2023any,du2023compositional,zhu2023learning,dutta2025co3}, as well as VLM-based feedback to better satisfy complex prompts \cite{wen2023improving}.

Multi-concept customization stresses compositional behavior when combining user-specific concepts \cite{ruiz2023dreambooth,kumari2023multi,liu2023cones2,jiang2025mc2,ding2024freecustom,kong2024omg,patashnik2025nested}.
In parallel, benchmarks and evaluator studies aim to measure compositional fidelity and diagnose failure modes at scale, often using pretrained VLMs as automated evaluators \cite{huang2023t2i,do2025vsc,wan2025compalign,hua2025mmig,kasaei2025evaluating,said2025deconstructing}.
Building on these advances in compositional generation and evaluation, we take a data-centric perspective and investigate an upstream driver of systematic dominance: concept-level visual homogeneity in training images.

\section{The Dominant-vs-Dominated Phenomenon}
\label{sec:dataset}

\subsection{Phenomenon Definition}
\label{subsec:dvd}

Prior work studies memorization---reproducing near-identical training examples from duplicated image--text pairs \cite{carlini2023extracting, wen2024detectingexplainingmitigatingmemorization, somepalli2023understanding}---and methods that improve multi-concept compositional generation \cite{hertz2023prompttoprompt, ding2024freecustom, brooks2023instructpix2pix, patashnik2025nested}.
Here we connect these perspectives by examining an upstream data property—concept-level visual homogeneity in training images—that can induce systematic visual dominance in multi-concept generation.

We define the DvD phenomenon as cases where one concept (the \emph{dominant} concept) consistently visually dominates the generation, while the other (the \emph{dominated} concept) is strongly suppressed and rarely represented.

\subsubsection{Illustrative example.}
As illustrated in \cref{fig:dvd_example}, the prompt ``Neuschwanstein Castle coaster'' demonstrates this phenomenon: across multiple random seeds, the distinctive architecture of Neuschwanstein Castle dominates the generation while the coaster concept is suppressed.
This pattern persists across SD 1.4 and SD 2.1 and is also observed in high-DvD newer-model examples, indicating that DvD is not simply a model-specific artifact.

\subsubsection{Hypothesis: training-image visual homogeneity.}
\begin{figure}[t]
    \centering

    \begin{minipage}{0.49\linewidth}
        \centering
        \includegraphics[width=\linewidth]{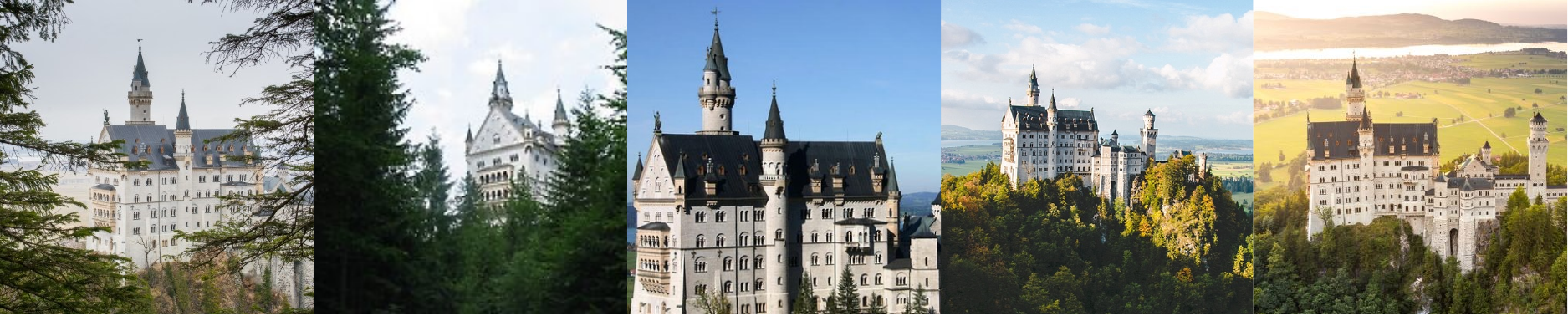}
        \small (a) Neuschwanstein Castle
    \end{minipage}
    \hfill
    \begin{minipage}{0.49\linewidth}
        \centering
        \includegraphics[width=\linewidth]{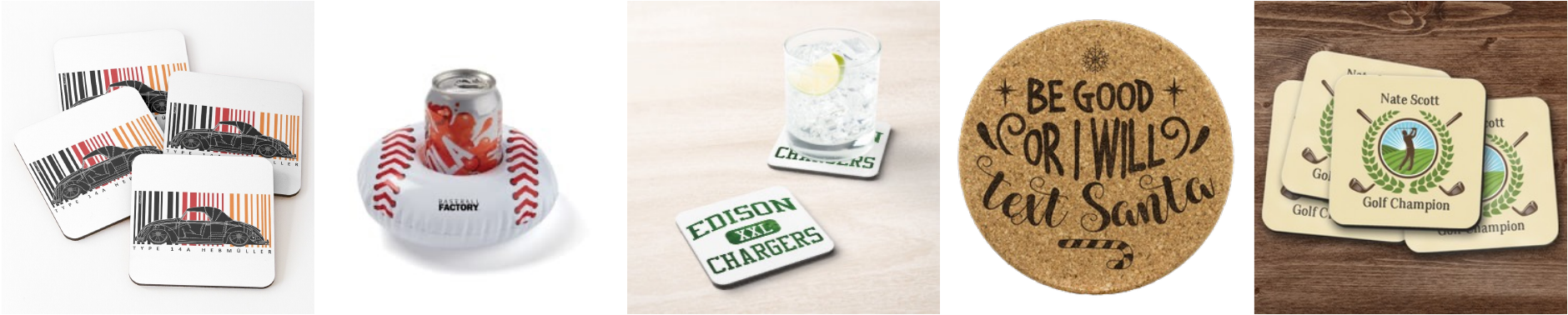}
        \small (b) Coaster
    \end{minipage}

    \vspace{-0.5em}
    \caption{Training data examples from LAION \cite{schuhmann2022laion}: Neuschwanstein Castle exhibits minimal visual variation, while coasters appear in diverse forms and contexts.}
    \vspace{-1.5em}
    \label{fig:training_data}
\end{figure}

We hypothesize that DvD stems from the visual homogeneity (i.e., limited variation) of concept-specific training images: concepts learned from visually homogeneous images may form relatively rigid visual priors that dominate others learned from more diverse images.
To investigate this, we examine training images from the LAION dataset \cite{schuhmann2022laion} for both concepts (\cref{fig:training_data}).
As shown in \cref{fig:training_data}a, Neuschwanstein Castle, being a unique landmark, appears with highly consistent visual features---the iconic white facade, pointed towers, and alpine setting remain nearly identical across all training images.
In contrast, \cref{fig:training_data}b reveals that coasters exist in diverse forms: round plates, square tiles, and decorative pieces with various colors, patterns, and materials.
These differences in training-image diversity can lead diffusion models to develop internal visual representations with different levels of flexibility.

\subsection{Quantification Metric}
\label{subsec:dvd_score}

To quantify dominance, we define the \emph{DvD Score} based on whether each concept is visually present in the generated image.
We focus on two-concept prompts for a direct dominant--dominated comparison.
Each concept is evaluated through $N$ binary questions using the VLM evaluator Qwen2.5-VL \cite{bai2025qwen2}.
Let $C_1$ and $C_2$ denote the numbers of ``Yes'' responses for the two concepts, respectively.
The \emph{DvD Score} is defined as:
\vspace{-0.5em}
\begin{equation}
\text{\emph{DvD Score}} = \frac{C_1 \times (N - C_2)}{N^2} \times 100.
\end{equation}

This metric ranges from $0$ to $100$, with higher values indicating stronger dominance.
We set $N=5$ with concept-type-specific questions (e.g., for artists: ``Is this image painted in the artistic style of Van Gogh?'').
Appendix~S1.2 provides the full set of questions and per-image response examples.

\subsection{DominanceBench}
\label{subsec:db}
To systematically investigate the mechanisms of DvD, we propose DominanceBench, a benchmark of 300 two-concept prompts collected from the LAION dataset \cite{schuhmann2022laion}, on which SD was trained.
We focus on prompts that pair low-diversity concepts (artists, landmarks, and characters) with everyday objects; the full concept lists are provided in Appendix~S1.1.
A LAION-based visual-homogeneity analysis in Appendix~S1.1 supports this category-level grouping.
\vspace{0.1cm}

\noindent
\begin{minipage}[t]{0.54\textwidth}
  \vspace{0pt}
  \small
  \hrule
  \vspace{0.25em}
  \refstepcounter{algorithm}
  \textbf{Algorithm \thealgorithm} DominanceBench DvD-Score-Based Prompt Filtering
  \label{alg:dominancebench_filtering}
  \vspace{0.25em}
  \hrule
  \vspace{0.35em}
  \begin{algorithmic}[1]
    \STATE Collect candidate prompts $p$ from LAION (20--50 chars) that contain a low-diversity concept and an object.
    \FOR{each prompt $p$}
      \STATE Generate 10 images $\{I_1,\ldots,I_{10}\}$ with SD~1.4 (different seeds).
      \STATE Compute DvD Score for each image $I_i$.
      \STATE Count $n = |\{i : \text{DvD Score}(I_i) > 36\}|$.
      \IF{$n \ge 7$}
        \STATE Add $p$ to DominanceBench.
      \ENDIF
    \ENDFOR
  \end{algorithmic}
  \hrule
\end{minipage}
\hfill
\begin{minipage}[t]{0.44\textwidth}
  \vspace{0pt}
  \centering
  \IfFileExists{plots/dvd_score_comparison_5models_with_new_arch.pdf}{%
    \includegraphics[width=\linewidth,trim=8pt 30pt 6pt 34pt,clip]{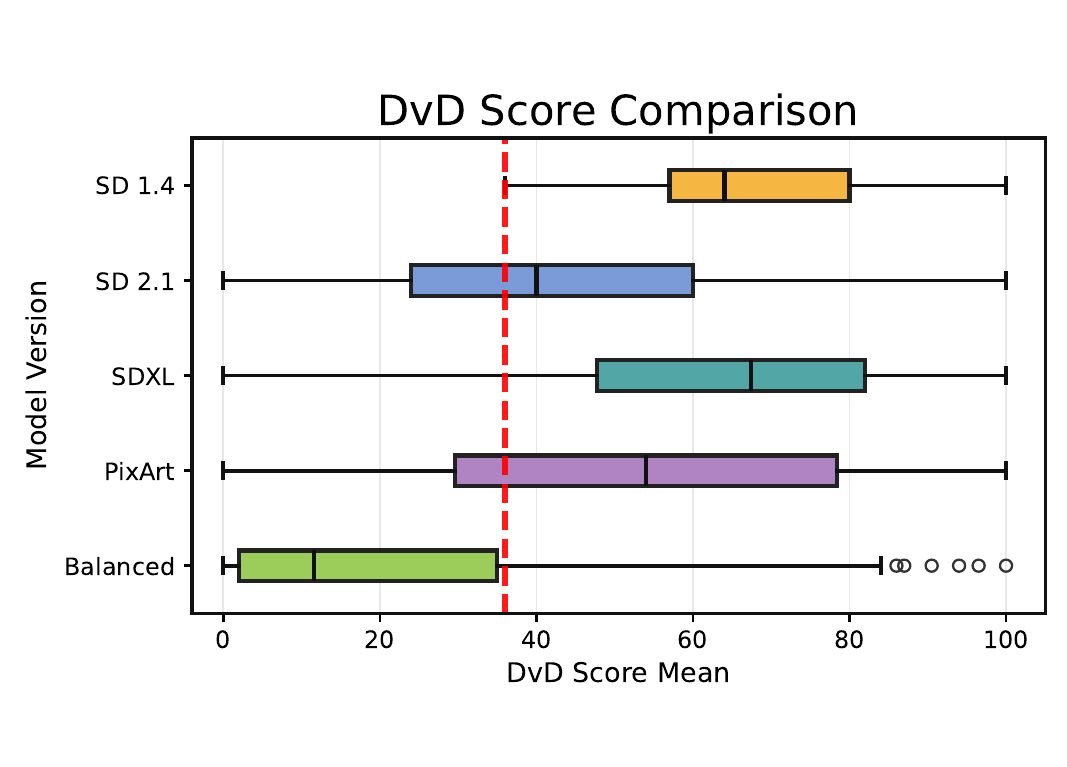}
  }{%
    \includegraphics[width=\linewidth]{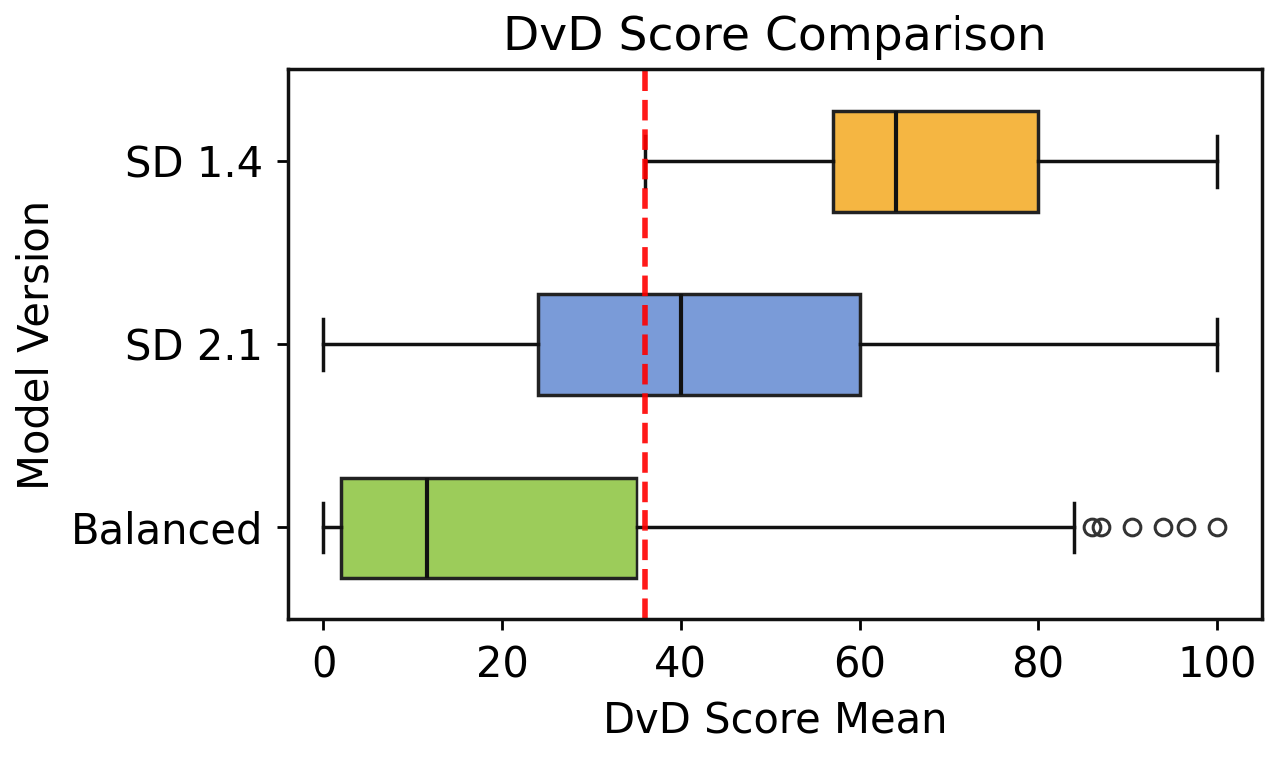}
  }
  \vspace{-1.0em}
  \captionsetup{aboveskip=0pt,belowskip=0pt}
  \captionof{figure}{10-seed mean DvD Scores for DominanceBench across SD 1.4, SD 2.1, SDXL, and PixArt, with balanced prompts for reference. Dashed line: threshold (36).}
  \label{fig:dvd_comparison}
\end{minipage}

\vspace{0.3cm}
More specifically, the 300 prompts in DominanceBench are selected by filtering prompts from LAION, as shown in \cref{alg:dominancebench_filtering}.
In short, if 7 out of 10 images generated from a candidate prompt with different random seeds show a DvD Score higher than 36, the prompt is considered as a DvD prompt and included in DominanceBench.
Here, the threshold is determined as follows. We build a separate set of 300 balanced two-concept prompts where both concepts successfully appear in the generated images (details in Appendix~S2.1). 
Then, the DvD Scores are measured for the 10 images per balanced prompt with different random seeds.
The third quartile of the score distribution is chosen as the threshold, as shown in \cref{fig:dvd_comparison}.
In the figure, it can be observed that the DominanceBench prompts have higher DvD Scores than the balanced prompts.
While the initial collection is performed using SD 1.4, a substantial portion of the DominanceBench prompts still exhibit high dominance in SD 2.1, as shown in \cref{fig:dvd_comparison}.
As an additional cross-architecture check, we evaluate DominanceBench on newer text-to-image models.
DvD remains pronounced in SDXL~\cite{podell2024sdxl} and PixArt~\cite{chen2024pixartalpha}, with 85.33\% and 67.33\% of prompts exceeding the threshold, respectively.
These high above-threshold rates suggest that DvD is not solely an artifact of SD 1.4 or SD 2.1.
We use SD 1.4 and SD 2.1 as the primary models for the mechanistic analyses in the remainder of the paper.

We assess the reliability of the VLM-based evaluation with human annotations on 300 generated images from DominanceBench and balanced prompts on SD~1.4 and SD~2.1, using yes/ambiguous/no concept-presence labels.
Across 17 annotators, the majority human label matches the Qwen2.5-VL label for 85.0\% of DominanceBench images and 90.0\% of balanced images.
Additional prompt-level robustness checks, including independently constructed prompts and threshold sweeps, are provided in Appendix~S1.3 and S1.4.

\vspace{-0.7em}
\section{Analysis}
\label{sec:analysis}
\vspace{-0.7em}
In this section, we analyze the causes and mechanisms of DvD.
We first validate our data-centric hypothesis via UNet fine-tuning experiments (\cref{subsec:toy_example}), then analyze cross-attention dynamics (\cref{subsec:cross_attention}), and finally conduct a head ablation study (\cref{subsec:head_ablation}) to compare DvD with memorization.

\vspace{-0.3cm}
\subsection{Verifying the Role of Visual Homogeneity in Training Images}
\label{subsec:toy_example}

\begin{tcolorbox}[colback=teal!10, colframe=teal!80, title=\textbf{Takeaway 1}, left=2pt, right=2pt, top=2pt, bottom=2pt]
    \textbf{Lower visual diversity can lead to stronger dominance.}
    When a concept is learned from limited variations, its representation becomes overfit to specific visual patterns, causing it to dominate other concepts in multi-concept compositions.
\end{tcolorbox}

In \cref{subsec:dvd}, we hypothesized that DvD stems from the visual homogeneity (i.e., limited variation) of concept-specific training images.
We test this hypothesis through controlled fine-tuning experiments that systematically vary within-concept visual diversity.

\vspace{-0.3em}
\subsubsection{Experimental Setup.}
We follow the DreamBooth setup for learning new concepts in SD fine-tuning \cite{ruiz2023dreambooth}.
Specifically, we fine-tune three separate models to learn ``sks castle'', ``sks mountain'', and ``sks arena'', each defined by pairing the identifier token ``sks'' with a class word.
We choose ``sks'' as a token without a meaningful semantic prior, helping isolate concept learning from pre-existing semantic associations in the text encoder.
To obtain concept-specific images with controllable diversity, we retrieve LAION images \cite{schuhmann2022laion} corresponding to three specific landmarks for castle, mountain, and arena, respectively---Neuschwanstein Castle, Machu Picchu, and Colosseum---whose images are visually highly similar, making it straightforward to construct highly homogeneous training sets.
We use these sites as the visual targets for the three class words while keeping the training prompt generic (to avoid name-specific priors).

We fine-tune only the UNet (keeping the text encoder fixed) of SD 1.4 and SD 2.1 to learn these concepts.
This setting provides a controlled way to study how training-image diversity affects concept learning in SD fine-tuning.
We train all models for 100 epochs with Adam~\cite{kingma2015adam}, using a learning rate of $1\times10^{-5}$.
We use LAION captions only for retrieval; to construct image--text pairs for fine-tuning, we generate a caption for each selected image using BLIP-2 \cite{li2023blip}.

\vspace{-0.7em}
\subsubsection{Diversity Control.}
We construct five 120-image training sets $\mathcal{D}_k$ ($k=1,\ldots,5$), where a smaller $k$ indicates a more visually homogeneous image set.

\begin{figure*}[t]
    \centering
    \begin{subfigure}[t]{0.47\linewidth}
        \centering
        \includegraphics[width=\linewidth]{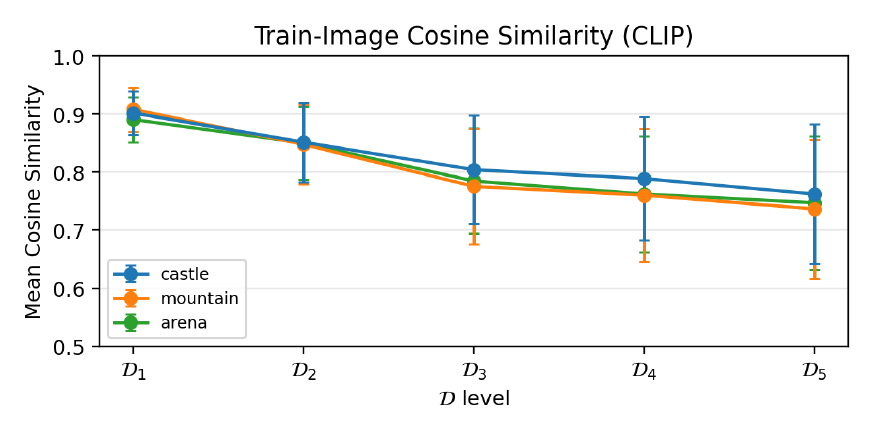}
        \caption{CLIP}
        \label{fig:diversity_control_clip}
    \end{subfigure}
    \hfill
    \begin{subfigure}[t]{0.47\linewidth}
        \centering
        \includegraphics[width=\linewidth]{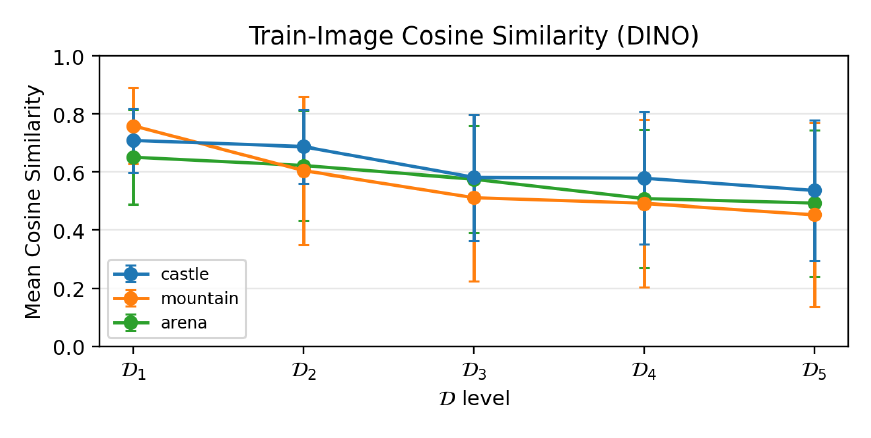}
        \caption{DINOv2}
        \label{fig:diversity_control_dino}
    \end{subfigure}
    \vspace{-0.7em}
    \caption{Diversity control and validation for the constructed training sets. We show mean intra-set cosine similarity (mean $\pm$ std) across homogeneity configurations $\mathcal{D}_k$ ($k=1,\ldots,5$), measured using CLIP (used to construct the sets) and DINOv2 (used for validation). Higher similarity indicates more visually homogeneous training images.}
    \label{fig:diversity_control}
    \vspace{-1.5em}
\end{figure*}

For each target concept, we retrieve images whose LAION captions contain the corresponding proper name.
We then construct each $\mathcal{D}_k$ by grouping and sampling images based on a cosine similarity threshold $\tau_k$ of CLIP \cite{radford2021learning} embeddings with $(\tau_1,\dots,\tau_5)=(0.95, 0.90, 0.85, 0.80, 0.75)$, where a higher $\tau_k$ yields a more visually homogeneous set (\cref{fig:diversity_control_clip}).

The controlled diversity ordering based on CLIP is additionally verified using DINOv2 \cite{oquab2023dinov2} embeddings. \cref{fig:diversity_control_dino} confirms that the overall diversity trend is consistent in both embeddings, supporting the reliability of the constructed training sets.

\vspace{-0.2cm}
\subsubsection{Evaluation.}
We evaluate dominance on 100 two-concept prompts that pair a learned concept (e.g., ``sks castle'') with diverse partner concepts.
We first construct 50 base prompts where the learned concept appears first, followed by the partner concept. Then, in order to eliminate concept-order effects, a flipped-order version of each base prompt is obtained, yielding 100 prompts in total.
For each prompt, we generate 10 images per fine-tuned model variant using different random seeds, compute the DvD Score (\cref{subsec:dvd_score}), and report per-prompt mean scores, treating the learned concept as $C_1$.
Example evaluation prompts are provided in Appendix~S3.1.

\vspace{-0.5em}
\subsubsection{Results.}
\cref{fig:toy_example_combined} summarizes the effect of diversity control for the learned concepts.
For both SD 1.4 and SD 2.1, the distributions of per-prompt mean DvD Scores tend to shift toward higher values as within-concept diversity decreases (smaller $k$ in $\mathcal{D}_k$).

\begin{figure*}[!t]
    \centering
    \begin{subfigure}[t]{0.3\linewidth}
        \centering
        \includegraphics[width=\linewidth]{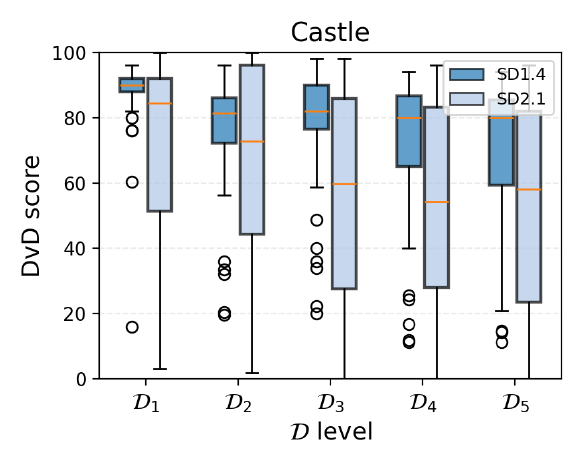}
        \caption{Castle}
    \end{subfigure}
    \hfill
    \begin{subfigure}[t]{0.30\linewidth}
        \centering
        \includegraphics[width=\linewidth]{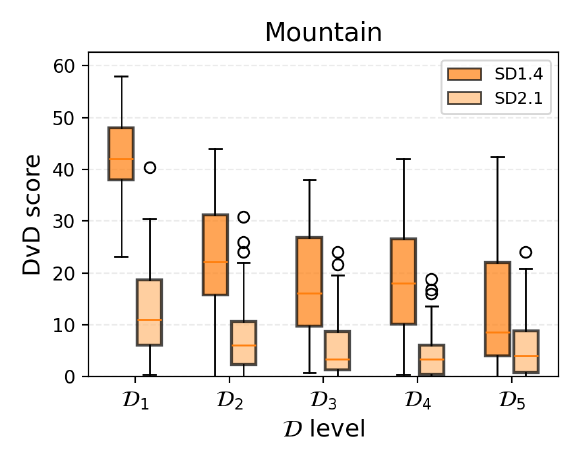}
        \caption{Mountain}
    \end{subfigure}
    \hfill
    \begin{subfigure}[t]{0.30\linewidth}
        \centering
        \includegraphics[width=\linewidth]{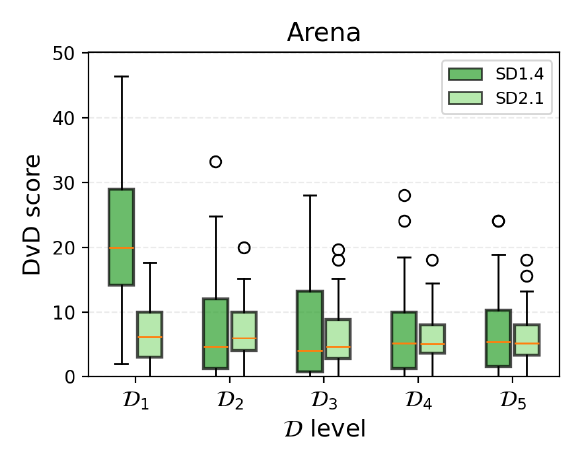}
        \caption{Arena}
    \end{subfigure}

    \vspace{-0.7em}
    \caption{Effect of visual diversity control on dominance. We show the distributions of per-prompt mean DvD Scores across different homogeneity configurations ($\mathcal{D}_1$--$\mathcal{D}_5$) for the three learned concepts.}
    \vspace{-1.5em}
    \label{fig:toy_example_combined}
\end{figure*}

Collectively, these controlled experiments show that, in this concept-learning setting, visual homogeneity in training images can increase dominance in multi-concept generation (higher DvD Scores).
This suggests that learning a concept from limited variations can yield overly rigid visual priors that suppress other concepts.
A supplementary re-evaluation with InternVL2.5-8B, using the same generated images, prompts, and concept-presence questions, shows the same qualitative trend; details are provided in Appendix~S3.3.
As a qualitative illustration, \cref{fig:toy_example_examples} shows generated examples across homogeneity configurations for each concept.
\vspace{-0.5em}

\begin{figure*}[!t]
    \centering
    \includegraphics[width=0.8\linewidth]{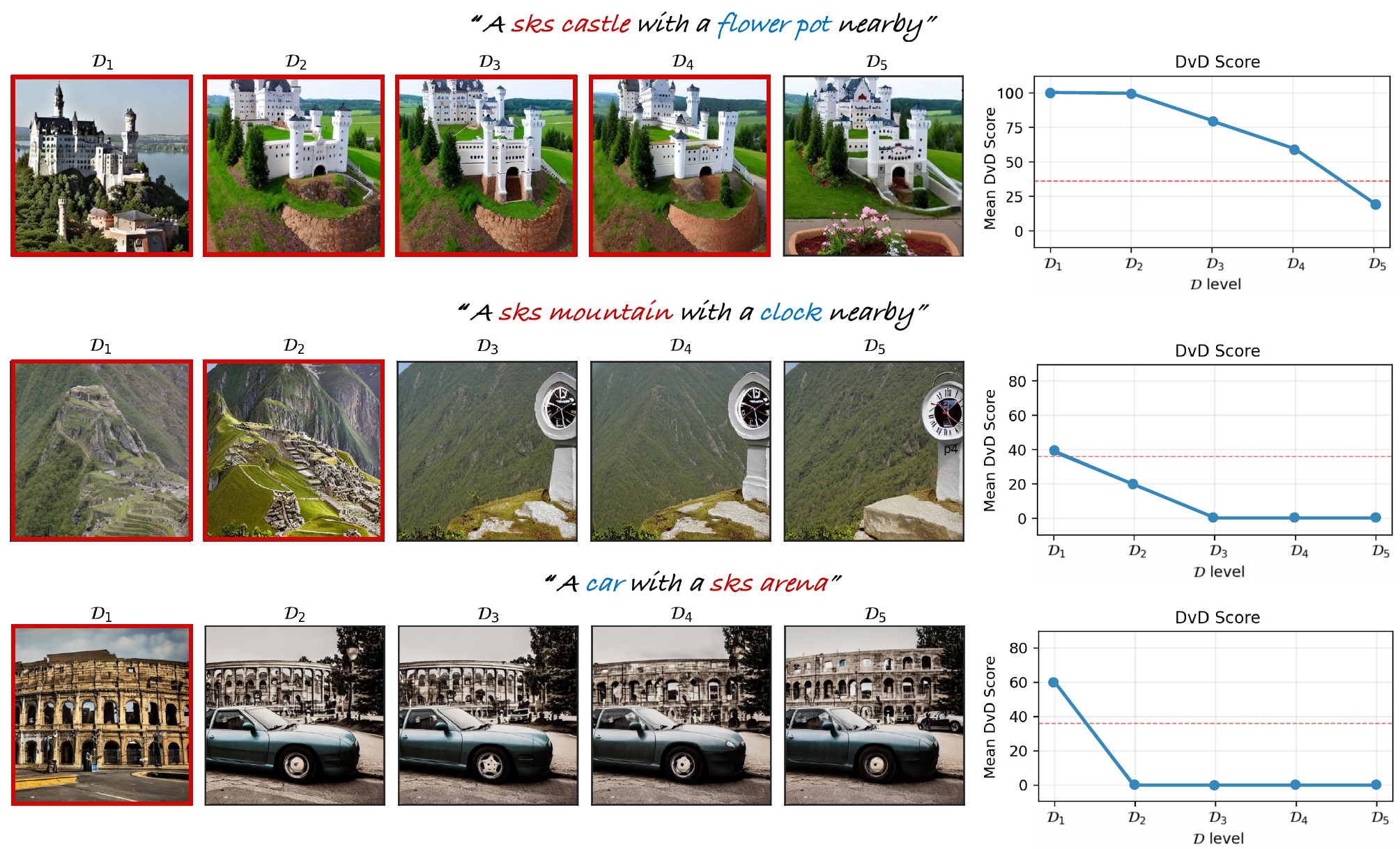}
    \vspace{-0.7em}
    \caption{Representative generation examples across homogeneity configurations for the three learned concepts. Images outlined in red indicate DvD cases.}
    \vspace{-1.5em}
    \label{fig:toy_example_examples}
\end{figure*}

\subsection{Cross-Attention Dynamics in DvD}
\label{subsec:cross_attention}

DvD can be viewed as a competition between concept word tokens in a multi-concept prompt, where one token's conditioning influence often dominates and suppresses the others.
Since prompt information is injected into the UNet primarily through cross-attention, we analyze cross-attention patterns in DominanceBench prompts to localize where attention becomes imbalanced and how this imbalance evolves through denoising timesteps.
We first fix the denoising timestep ($t=50$) and scan UNet layers to identify where attention concentrates, then track how attention to the dominant and dominated tokens changes over subsequent timesteps in these layers.
\vspace{-1.7em}

\subsubsection{Layer-wise Attention Focus at the First Denoising Step.}
\label{subsubsec:focus_score}

\par\vspace{0.5em}
\noindent

\begin{tcolorbox}[
    colback=teal!10,
    colframe=teal!80,
    title=\textbf{Takeaway 2},
    left=2pt,
    right=2pt,
    top=2pt,
    bottom=2pt
    ]
    The dominant token in a DvD prompt tends to receive the maximum attention in low-resolution blocks already at the first denoising step.
\end{tcolorbox}

To quantify how strongly the model focuses on specific tokens, we define a \textit{focus score}:
\begin{equation}
\label{eq:focus_score}
\text{Focus}^{(\ell, t)} = \frac{\max_i a_i^{(\ell, t)} - \bar{a}_{\text{others}}^{(\ell, t)}}{H(\mathbf{a}^{(\ell, t)}) / \log_2 N + \epsilon}
\end{equation}
where $\mathbf{a}^{(\ell, t)} = (a_1^{(\ell, t)}, \ldots, a_N^{(\ell, t)})$ represents the cross-attention weights over $N$ prompt tokens at layer $\ell$ and timestep $t$ (averaged across all spatial locations and attention heads), $\max_i a_i^{(\ell, t)}$ is the maximum attention weight, $\bar{a}_{\text{others}}^{(\ell, t)}$ is the mean of all other attention weights, $H(\mathbf{a}^{(\ell, t)})$ is the entropy of the attention distribution, and $\epsilon$ is a small constant for numerical stability.
We provide additional motivation for this entropy-based normalization in Appendix~S5.

Intuitively, the focus score captures whether cross-attention forms a sharp peak on a single token or is distributed evenly across the prompt. Higher values indicate stronger single-token concentration.

\vspace{-0.7em}
\subsubsection{Experimental Setup.}
We compute focus scores in all UNet layers during the first denoising step ($t=50$).
Following the UNet ordering of cross-attention modules, layers 1--6 correspond to the downsampling blocks, layer 7 to the mid block, and layers 8--16 to the upsampling blocks.
In particular, layers 5--10 comprise the low-resolution stage.

\begin{figure*}[t]
    \centering
    \begin{subfigure}[t]{0.49\textwidth}
        \centering
        \includegraphics[width=\linewidth]{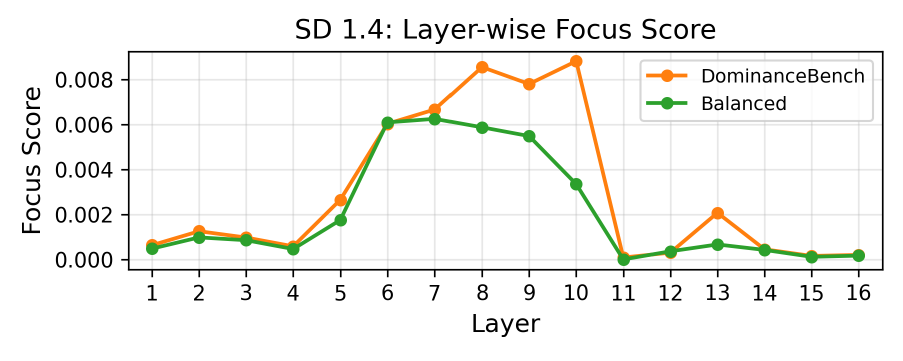}
        \caption{SD 1.4: Focus score}
        \label{fig:focus_sd14}
    \end{subfigure}
    \hfill
    \begin{subfigure}[t]{0.49\textwidth}
        \centering
        \includegraphics[width=\linewidth]{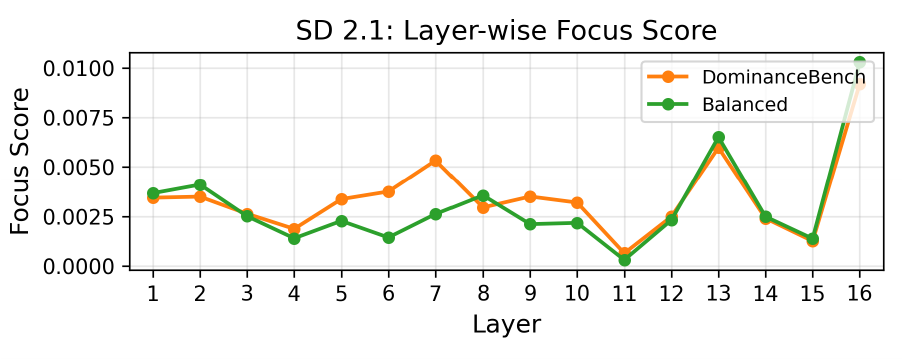}
        \caption{SD 2.1: Focus score}
        \label{fig:focus_sd21}
    \end{subfigure}
    \vspace{-0.7em}
    \caption{Focus scores across UNet layers at the first denoising step ($t=50$) for the DominanceBench prompts and the balanced prompts.}
    \label{fig:layerwise_focus}
    \vspace{-1.0em}
\end{figure*}

\vspace{-0.7em}
\subsubsection{Results.}
In \cref{fig:layerwise_focus}, the DominanceBench prompts show a distinctive pattern in focus scores (i.e., significantly higher focus scores) in the low-resolution stage (layers 5--10) compared to the balanced prompts. This indicates that cross-attention becomes sharply peaked already at the first denoising step, i.e., one token receives excessive attention weights. 

We further examine which specific concept token is emphasized.
Since layers 8--10 show a particularly large focus score gap between DominanceBench and balanced prompts in both SD 1.4 and SD 2.1, we identify the maximally attended token in these layers.
In SD 1.4, the dominant concept token receives the maximum attention in 253 out of 300 DominanceBench prompts (84.33\%). Similarly in SD 2.1, the dominant concept token is the most frequently maximally attended token (162 out of 300 prompts; 54.00\%).

This confirms that the influence of the dominant concept token becomes amplified through the cross-attention operation in the low-resolution stage at the first denoising step. In what follows, we investigate the temporal dynamics of the influences of the dominant and dominated concepts over the subsequent denoising steps.


\vspace{-0.7em}
\subsubsection{Temporal Dynamics of Cross-Attention.}
\label{subsubsec:temporal_analysis}

\par\vspace{0.5em}
\noindent

\begin{tcolorbox}[
    colback=teal!10,
    colframe=teal!80,
    title=\textbf{Takeaway 3},
    left=2pt,
    right=2pt,
    top=2pt,
    bottom=2pt
]
\textbf{Dominated concepts rapidly lose attention in early denoising timesteps.}
Within the first few denoising steps, attention to the dominated token fades, while the dominant token retains a clear peak.
\end{tcolorbox}

\begin{figure*}[t]
    \centering
    \begin{subfigure}[t]{0.32\textwidth}
        \centering
        \includegraphics[width=\linewidth]{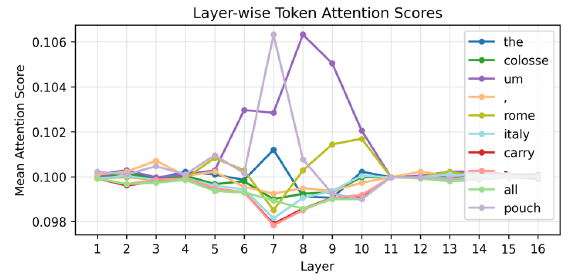}
        \captionsetup{labelsep=newline}
        \caption{Layer-wise ($t=50$)}
        \label{fig:attn_graph}
    \end{subfigure}
    \hfill
    \begin{subfigure}[t]{0.35\textwidth}
        \centering
        \includegraphics[width=\linewidth]{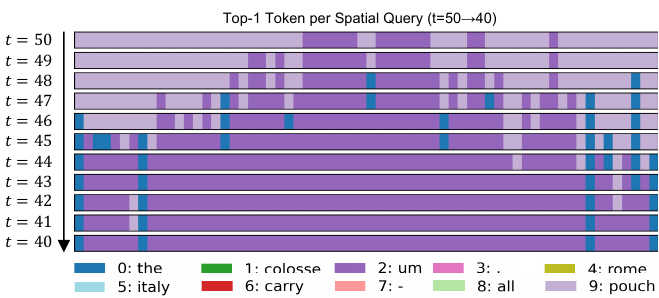}
        \captionsetup{labelsep=newline}
        \caption{Top-1 ($t=50\rightarrow40$)}
        \label{fig:attn_top1_timeline}
    \end{subfigure}
    \hfill
    \begin{subfigure}[t]{0.125\textwidth}
        \centering
        \includegraphics[width=\linewidth]{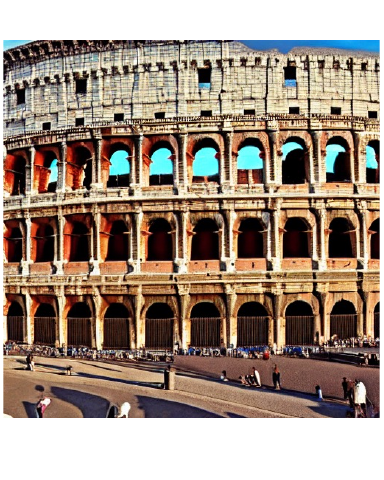}
        \captionsetup{format=plain,justification=centering,singlelinecheck=false,labelsep=newline}
        \caption{DvD}
        \label{fig:attn_dvd}
    \end{subfigure}
    \hfill
    \begin{subfigure}[t]{0.125\textwidth}
        \centering
        \includegraphics[width=\linewidth]{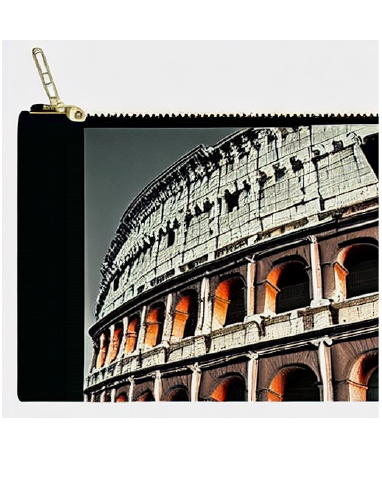}
        \captionsetup{format=plain,justification=centering,singlelinecheck=false,labelsep=newline}
        \caption{Balanced}
        \label{fig:attn_balanced}
    \end{subfigure}

    \vspace{-0.7em}
    \caption{Example cross-attention patterns for a DominanceBench prompt. (a) Layer-wise token attention at the first denoising step ($t=50$). (b) Top-1 token maps across spatial locations from $t=50$ to $t=40$; regions where the dominated token (``pouch'') is top-1 rapidly diminish. (c) DvD outcome (dominated concept missing). (d) Balanced outcome (both concepts present).}
    \label{fig:layer_attention_example}
\end{figure*}

In some DvD prompts, the dominated token still receives substantial attention at $t=50$, yet it can fail to appear in the final image.
\cref{fig:layer_attention_example} shows one such case (``The Colosseum Rome Italy Carry-all Pouch''): at $t=50$, the token ``pouch'' is strongly attended in the mid block (\cref{fig:attn_graph}).
Yet the spatial locations where it is the top-1 token (highest-attended token per spatial location) quickly shrink from $t=50$ to $t=40$ (\cref{fig:attn_top1_timeline}), leading to a DvD outcome (\cref{fig:attn_dvd}) rather than balanced generation (\cref{fig:attn_balanced}).
To characterize when this reduction occurs, we track how attention to the dominant and dominated tokens evolves over early denoising timesteps.

\vspace{-1.5em}
\subsubsection{Experimental Setup.}
\begin{figure*}[t]
    \centering
    \begin{subfigure}[t]{0.45\textwidth}
        \centering
        \includegraphics[width=\linewidth]{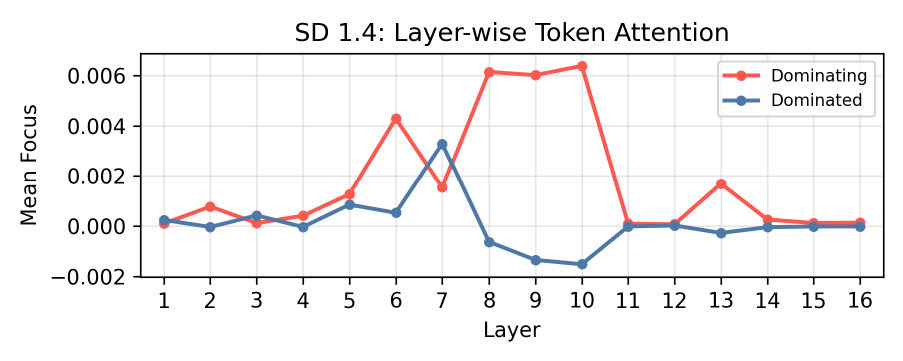}
        \caption{SD 1.4}
        \label{fig:token_focus_sd14}
    \end{subfigure}
    \hfill
    \begin{subfigure}[t]{0.45\textwidth}
        \centering
        \includegraphics[width=\linewidth]{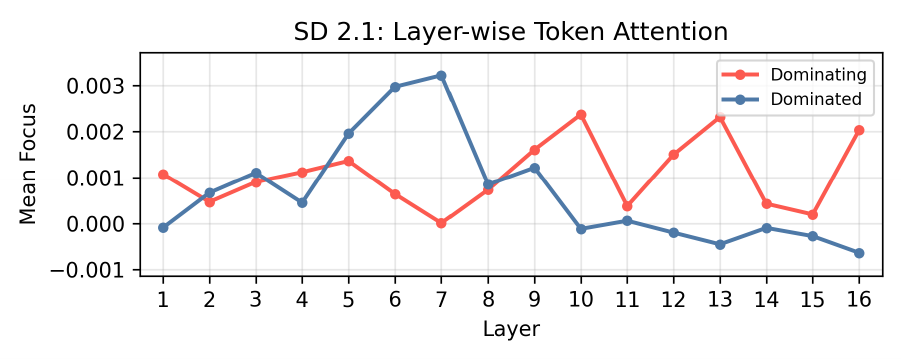}
        \caption{SD 2.1}
        \label{fig:token_focus_sd21}
    \end{subfigure}
    \vspace{-0.7em}
    \caption{Layer-wise token attention at the first denoising step ($t=50$) for the dominant and dominated tokens (averaged over prompts).}
    \label{fig:first_step_token_profiles}
\end{figure*}

Using the dominant/dominated token labels provided by DominanceBench, we compute layer-wise attention at the first denoising step ($t=50$) to decide where to track each token (\cref{fig:first_step_token_profiles}).
Across both models, the dominant token shows the peak in the first upsampling blocks (layers 8--10), while the dominated token peaks in the mid block (layer 7).
We therefore track the dominant token in layers 8--10 and the dominated token in layer 7 over early denoising timesteps.

To quantify the temporal evolution of attention for these tokens, we define the attention deviation for token $i$ at timestep $t$ as $\alpha_i^{(\ell, t)} = a_i^{(\ell, t)} - \bar{a}_{\text{others}}^{(\ell, t)}$ (with the same averaging across spatial locations and heads as in \cref{eq:focus_score}).
Note that we use the raw attention deviation rather than the entropy-normalized focus score, since we track changes over timesteps within a single prompt (with fixed prompt length).
Then, the attention change is:
\vspace{-0.5em}
\begin{equation}
\Delta \alpha_i^{(\ell, t)} = \alpha_i^{(\ell, t+1)} - \alpha_i^{(\ell, t)}
\end{equation}
In what follows, we average the attention changes over $\ell \in \{8,9,10\}$ for the dominant token and use $\ell=7$ for dominated tokens.
Negative $\Delta\alpha$ indicates decreasing concentration on that token.


\begin{figure*}[t]
    \centering
    \begin{subfigure}[b]{0.24\textwidth}
        \centering
        \includegraphics[width=\linewidth]{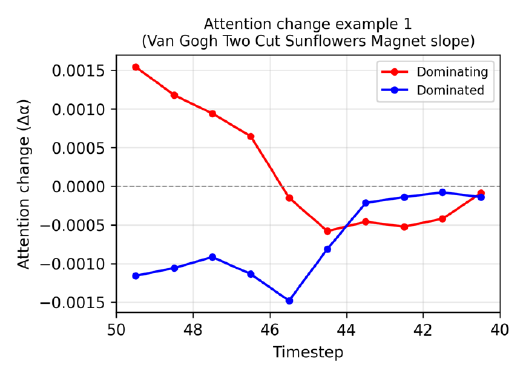}
        \caption{Example 1}
        \label{fig:slope_case_1}
    \end{subfigure}
    \hfill
    \begin{subfigure}[b]{0.24\textwidth}
        \centering
        \includegraphics[width=\linewidth]{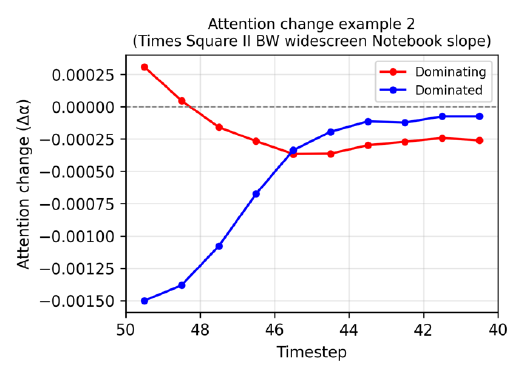}
        \caption{Example 2}
        \label{fig:slope_case_2}
    \end{subfigure}
    \hfill
    \begin{subfigure}[b]{0.24\textwidth}
        \centering
        \includegraphics[width=\linewidth]{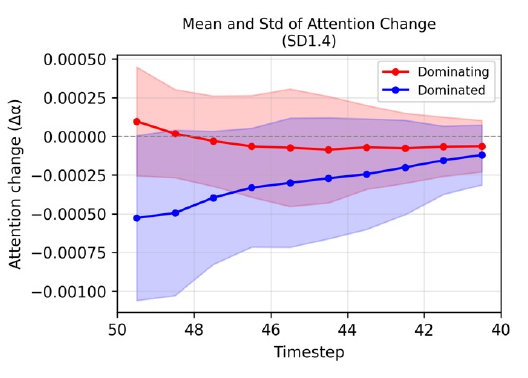}
        \caption{All (SD 1.4)}
        \label{fig:slope_mean_sd14}
    \end{subfigure}
    \hfill
    \begin{subfigure}[b]{0.24\textwidth}
        \centering
        \includegraphics[width=\linewidth]{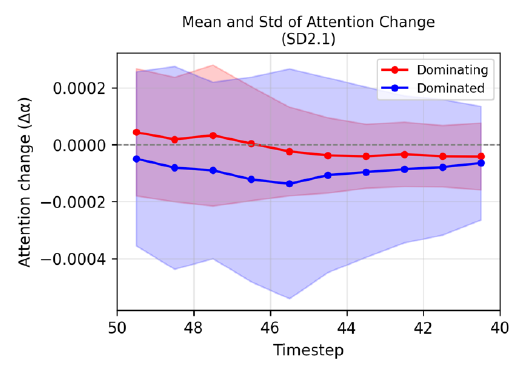}
        \caption{All (SD 2.1)}
        \label{fig:slope_mean_sd21}
    \end{subfigure}

    \vspace{-0.8em}
    \caption{Attention change ($\Delta\alpha$) across timestep intervals. (a,b) Two SD 1.4 examples. (c,d) Aggregated trends over all DominanceBench prompts on SD 1.4 and SD 2.1, respectively. The dominated token (blue) tends to show negative changes from the earliest timesteps, while the dominant token (red) maintains positive or near-zero changes initially.}
    \label{fig:temporal_slope}
    \vspace{-1.5em}
\end{figure*}

\subsubsection{Results.}
\cref{fig:temporal_slope} summarizes attention changes ($\Delta\alpha$) across early timestep intervals.
Individual prompts (Fig.~\ref{fig:slope_case_1},\ref{fig:slope_case_2}) show that the dominated token already has negative $\Delta\alpha$ values in the earliest timestep interval (50--40), indicating rapid attention decrease.
The prompt-averaged curves over all DominanceBench prompts (Fig.~\ref{fig:slope_mean_sd14},\ref{fig:slope_mean_sd21}) show a similar pattern on both SD 1.4 and SD 2.1: the dominated token tends to stay below zero, while the dominant token remains positive or near-zero initially.
This suggests that the imbalance is set up within the first few timesteps, and the dominated token loses its conditioning effect as denoising progresses.

\subsection{Head Ablation Study}
\label{subsec:head_ablation}
\vspace{-0.2em}
\begin{tcolorbox}[
    colback=teal!10,
    colframe=teal!80,
    title=\textbf{Takeaway 4},
    left=2pt,
    right=2pt,
    top=2pt,
    bottom=2pt
]
\textbf{At the head level, our ablation results suggest that DvD is more distributed across attention heads, whereas memorization tends to be localized to a small subset of heads.}
Accordingly, DvD is harder to mitigate by ablating only a few heads.
\end{tcolorbox}

Building on our cross-attention analysis, we move from layer-level patterns to individual attention heads---a practical unit of intervention within multi-head attention.
We aim to answer the following question: \emph{is DvD driven by a small set of critical heads, or does it emerge from distributed contributions across heads?}

To ground this head-level analysis, we use memorization as a reference point.
Neuron-level analyses suggest that memorization can be localized to a small number of internal units \cite{hintersdorf2024finding}.
Motivated by this, we use head ablation as a coarse probe to compare head-level localization between DvD and memorization, using 300 DvD prompts from DominanceBench and 500 memorized prompts identified in prior work \cite{carlini2023extracting}, respectively.

\vspace{-0.5cm}
\subsubsection{Head Ablation Procedure.}
To assess each head's contribution, we ablate target heads by scaling their attention logits (pre-softmax scores) with a small factor throughout denoising.
Formally, the ablated attention logit for head $h$ in layer $\ell$ is:
\begin{equation}
    \label{eq:head_ablation}
    \tilde{s}^{(\ell,h)} =
    \begin{cases}
    \varepsilon \cdot s^{(\ell,h)}, & \text{if } \ell=\ell^\star,\ h\in\mathcal{H}^\star, \\[3pt]
    s^{(\ell,h)}, & \text{otherwise},
    \end{cases}
\end{equation}
where $s^{(\ell,h)} \in \mathbb{R}^{P \times N}$ are the attention logits for head $h$ over $P$ spatial queries and $N$ text tokens (computed at each denoising step), $\ell^\star$ is the target layer, $\mathcal{H}^\star \subseteq \{1,\dots,H\}$ is the set of ablated heads, and $\varepsilon = 10^{-5}$ is a small scaling factor that effectively suppresses the head's influence.

\vspace{-0.7em}
\subsubsection{Outcome Classification.}
\begin{figure*}[t]
  \centering
  \begin{subfigure}[b]{0.45\linewidth}
    \includegraphics[width=\linewidth]{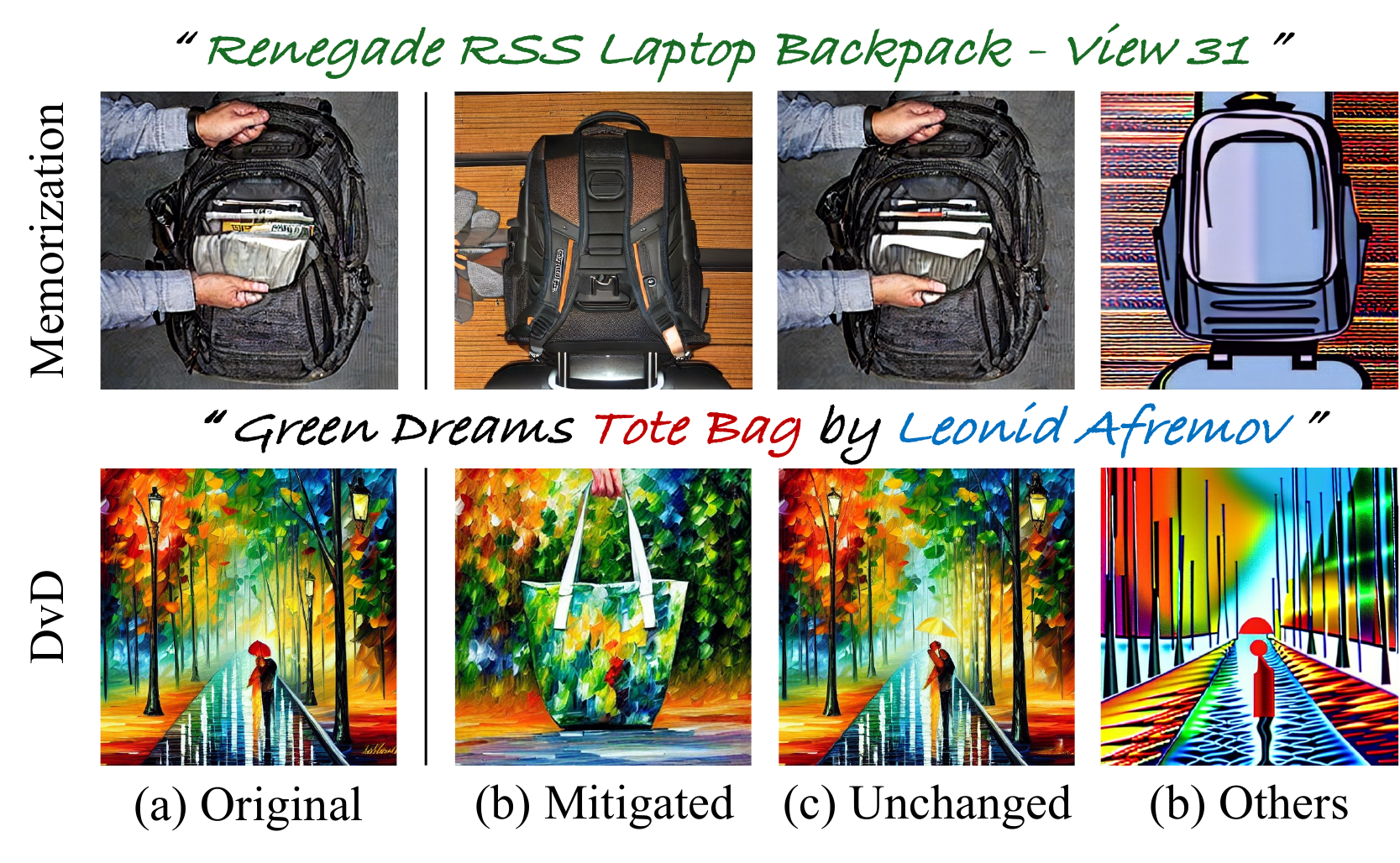}
  \end{subfigure}
  \hfill
  \begin{subfigure}[b]{0.45\linewidth}
    \includegraphics[width=\linewidth]{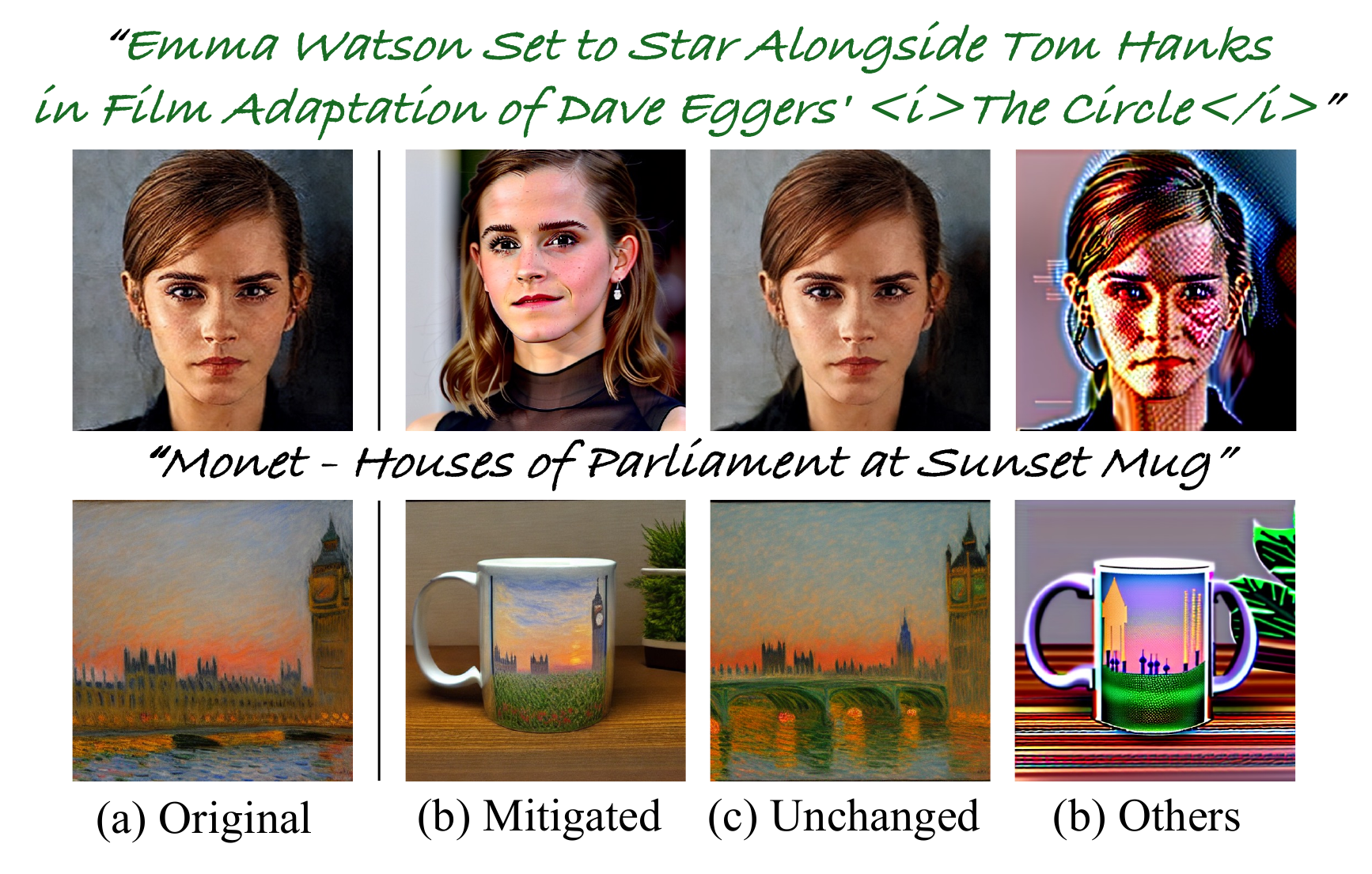}
  \end{subfigure}
  \vspace{-1.0em}
  \caption{Ablation outcomes. Each panel shows, left to right, the unablated generation, \textit{Mitigated}, \textit{Unchanged}, and \textit{Others} (degraded).}
  \vspace{0.6em}
  \label{fig:head_ablation_example}
\end{figure*}

We classify each generation result after ablation into three categories using Self-Supervised Descriptor for Image Copy Detection (SSCD) \cite{pizzi2022selfsuperviseddescriptorimagecopy}, Learned Perceptual Image Patch Similarity (LPIPS) \cite{zhang2018unreasonableeffectivenessdeepfeatures}, and (for DvD prompts) the DvD Score (\cref{subsec:dvd_score}), as illustrated in \cref{fig:head_ablation_example}.
SSCD measures similarity to training images (lower indicates less similarity / less copying in our usage).
LPIPS measures perceptual distance (higher indicates larger visual change).

\vspace{0.5em}

\vspace{-1.0em}

\begin{itemize}
  \item \textbf{Mitigated:} The target phenomenon (memorization or DvD) is successfully reduced.
    \begin{itemize}
    \item \emph{Memorization}: SSCD $< 0.5$ and LPIPS $> 0.6$
    \item \emph{DvD}: LPIPS $> 0.5$ and DvD Score $< 36$
    \end{itemize}
    
  \item \textbf{Unchanged:} The ablated output remains similar to the unablated generation and the target phenomenon persists.
    \begin{itemize}
    \item \emph{Memorization}: SSCD $\geq 0.5$ or LPIPS $\leq 0.6$
    \item \emph{DvD}: LPIPS $\leq 0.5$ or DvD Score $\geq 36$
    \end{itemize}
    
  \item \textbf{Others:} The generated image quality degrades substantially (e.g., severe artifacts or incoherent structure).
\end{itemize}

\vspace{-0.5cm}
\subsubsection{Single-Head Ablation.}
\label{subsubsec:single_head}

We first ablate individual heads by setting $|\mathcal{H}^\star| = 1$ in \cref{eq:head_ablation}.
For each prompt $p \in \mathcal{P}$ (where $\mathcal{P}$ contains 300 DominanceBench prompts or 500 memorization prompts), we test all heads in every layer (1--16), i.e., all $h \in \{1,\dots,H\}$ for every layer of the model.

\begin{figure*}[t]
  \centering
  \begin{subfigure}[b]{0.32\linewidth}
    \centering
    \includegraphics[width=\linewidth]{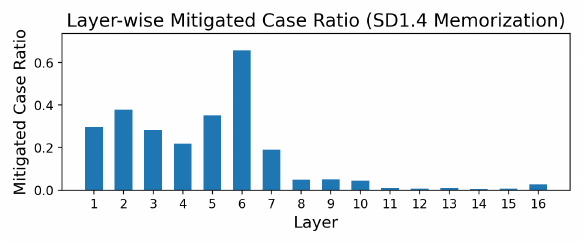}
    \caption{Memorization (SD 1.4)}
  \end{subfigure}
  \hfill
  \begin{subfigure}[b]{0.32\linewidth}
    \centering
    \includegraphics[width=\linewidth]{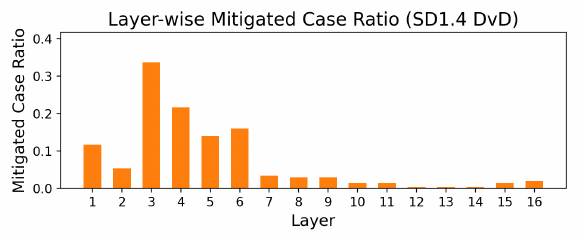}
    \caption{DvD (SD 1.4)}
  \end{subfigure}
  \hfill
  \begin{subfigure}[b]{0.32\linewidth}
    \centering
    \includegraphics[width=\linewidth]{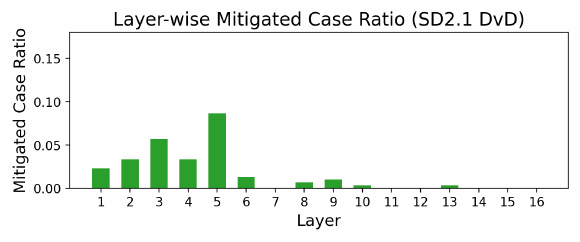}
    \caption{DvD (SD 2.1)}
  \end{subfigure}
  \vspace{-0.7em}
  \caption{Layer-wise ratio of mitigated cases in single-head ablation. For each layer, we report the proportion of the prompts that can be mitigated by ablating \emph{any} head in that layer. Mitigation effects concentrate primarily in layers 1--6.}
  \label{fig:single_head_layer_dist}
  \vspace{-1.0em}
\end{figure*}

We consider a prompt mitigated if ablating any head (in any layer) yields a \textit{Mitigated} outcome.
Under this criterion, single-head ablation mitigates 388 out of 500 memorization prompts on SD 1.4 (77.6\%), but only 136/300 (45.3\%) on SD 1.4 and 36/300 (12.0\%) on SD 2.1 for DvD.
As shown in \cref{fig:single_head_layer_dist}, mitigation concentrates in layers 1--6, i.e., the UNet downsampling blocks, with limited effects in higher layers.


\vspace{0.2cm}
\subsubsection{Multi-Head Ablation.}
\label{subsubsec:multi_head}

Next, we conduct multi-head ablation for $|\mathcal{H}^\star|=2$ and 3 in order to examine whether the DvD phenomenon is localized in a few heads or is distributed across a larger number of heads.
We focus on the downsampling blocks, where single-head ablation shows high mitigation rates (\cref{fig:single_head_layer_dist}).
As enumerating all head pairs/triplets quickly becomes expensive (a layer with $H$ heads has $\binom{H}{2}$ possible pairs and $\binom{H}{3}$ possible triplets), we instead randomly sample 10 head pairs ($|\mathcal{H}^\star|=2$) and 10 head triplets ($|\mathcal{H}^\star|=3$) per layer.
For each prompt $p$ and layer $\ell$, we ablate the heads in each sampled set simultaneously following \cref{eq:head_ablation} and classify the resulting images as \textit{Mitigated}, \textit{Unchanged}, or \textit{Others}.
We compute the outcome proportions over the sampled sets and average them over prompts to obtain the layer-wise outcome proportions (which sum to one).

\begin{figure*}[t]
  \centering

  \begin{subfigure}[b]{0.32\linewidth}
    \centering
    \includegraphics[width=\linewidth]{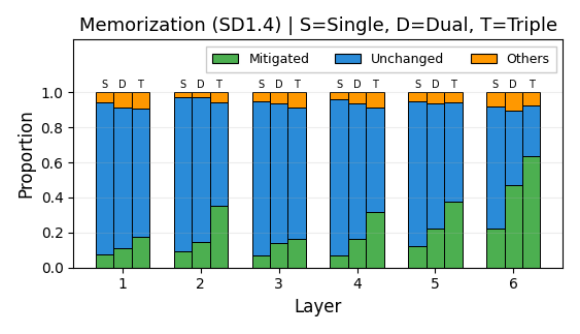}
    \caption{Memorization (SD 1.4)}
  \end{subfigure}
  \hfill
  \begin{subfigure}[b]{0.32\linewidth}
    \centering
    \includegraphics[width=\linewidth]{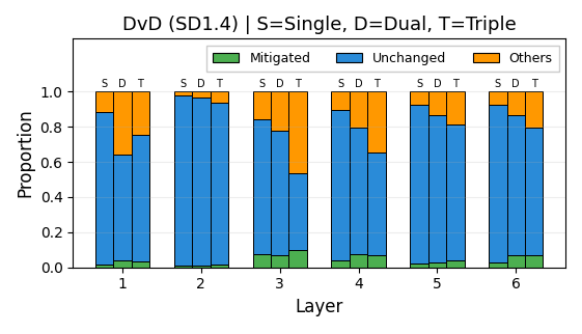}
    \caption{DvD (SD 1.4)}
  \end{subfigure}
  \hfill
  \begin{subfigure}[b]{0.32\linewidth}
    \centering
    \includegraphics[width=\linewidth]{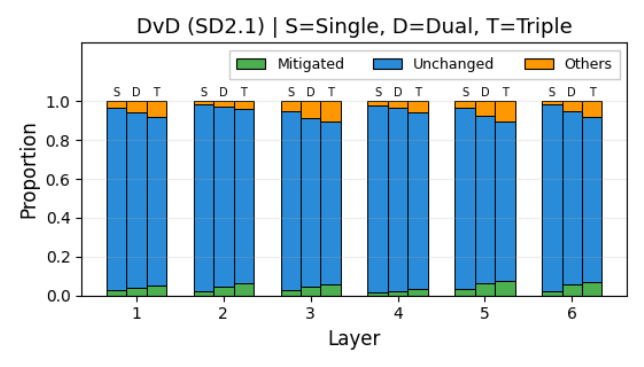}
    \caption{DvD (SD 2.1)}
  \end{subfigure}

  \vspace{-0.7em}
  \caption{Multi-head ablation outcome proportions across layers 1--6 as the ablation size increases (S=single, D=dual, T=triple).}
  \vspace{-1.0em}
  \label{fig:head_ablation_comparison}
\end{figure*}

\cref{fig:head_ablation_comparison} summarizes how outcome proportions change as we ablate more heads within the same layer.
For memorization (SD 1.4), increasing the ablation size mainly shifts outcomes from \textit{Unchanged} to \textit{Mitigated} while keeping \textit{Others} low, consistent with head-level localization.
For DvD, however, mitigation improves little as we increase the ablation size. On SD 1.4, outcomes increasingly shift to \textit{Others} rather than \textit{Mitigated}, meaning that ablation of more heads mainly causes degraded outputs with little increase in \textit{Mitigated}. On SD 2.1, most cases remain \textit{Unchanged} even under dual/triple ablations.
In sum, these patterns suggest that DvD is supported by contributions spread across many heads rather than by a small set of critical heads.

\section{Conclusion}
\label{sec:conclusion}
This work analyzed the DvD phenomenon in text-to-image diffusion models and introduced DominanceBench (300 prompts) for systematic study.
Our controlled fine-tuning experiments showed that visual homogeneity in concept-specific training images can increase dominance, suggesting that limited visual variation may induce overly rigid visual priors.
Mechanistically, DvD arises from sharply peaked cross-attention in low-resolution blocks at the first denoising step and an early decrease of the dominated token's conditioning influence.
Head ablations further suggested that, unlike memorization, DvD is supported by the contributions of many attention heads; ablating more heads tends to degrade the outputs without reliably reducing dominance.

These results point to a practical data-curation lever: monitoring and reducing concept-level visual redundancy (near-duplicates or highly homogeneous images) during dataset curation, and encouraging diverse reference sets in customization.
Finally, the proposed focus score provides a lightweight generation-time diagnostic at the first denoising step, enabling early detection and debugging (see Appendix~S8 for additional results).


\bibliographystyle{splncs04}
\bibliography{main}

\clearpage
\begin{center}
\Large\textbf{Dominant vs. Dominated: Concept-Level Generative Collapse in Diffusion Models}
\end{center}
\vspace{0.5em}

\begin{center}
\large{Supplementary Material}
\end{center}
\vspace{1.3em}

\renewcommand{\thesection}{S\arabic{section}}
\renewcommand{\thefigure}{S\arabic{figure}}
\renewcommand{\thetable}{S\arabic{table}}
\renewcommand{\theHsection}{supp.\arabic{section}}
\renewcommand{\theHsubsection}{supp.\arabic{section}.\arabic{subsection}}
\renewcommand{\theHsubsubsection}{supp.\arabic{section}.\arabic{subsection}.\arabic{subsubsection}}
\renewcommand{\theHfigure}{supp.\arabic{figure}}
\renewcommand{\theHtable}{supp.\arabic{table}}

\setcounter{section}{0} 
\setcounter{figure}{0}
\setcounter{table}{0}


\begin{center}
\large\textbf{Table of Contents}
\end{center}
\vspace{0.5em}

{%
\setlength{\parindent}{0pt}%
\setlength{\parskip}{1.0em}%
\textbf{\ref{sec:supp_dominancebench}. DominanceBench} \dotfill \pageref{sec:supp_dominancebench}\par
\makebox[1.5em][l]{}\ref{subsec:supp_training_data_diversity}. Training-Image Visual Homogeneity Analysis \dotfill \pageref{subsec:supp_training_data_diversity}\par
\makebox[1.5em][l]{}\ref{subsec:supp_prompt_collection_eval}. Prompt Collection and Evaluation Details \dotfill \pageref{subsec:supp_prompt_collection_eval}\par
\makebox[1.5em][l]{}\ref{subsec:supp_independent_prompt_set}. Independent Prompt Set \dotfill \pageref{subsec:supp_independent_prompt_set}\par
\makebox[1.5em][l]{}\ref{subsec:supp_threshold_robustness}. Threshold Robustness \dotfill \pageref{subsec:supp_threshold_robustness}\par

\textbf{\ref{sec:supp_reference_prompts}. Reference Prompts} \dotfill \pageref{sec:supp_reference_prompts}\par
\makebox[1.5em][l]{}\ref{subsec:supp_balanced_prompts}. Balanced Prompts \dotfill \pageref{subsec:supp_balanced_prompts}\par
\makebox[1.5em][l]{}\ref{subsec:supp_memorized_prompts}. Memorized Prompts \dotfill \pageref{subsec:supp_memorized_prompts}\par

\textbf{\ref{sec:supp_training}. Visual Homogeneity Experiments: Details and Extensions} \dotfill \pageref{sec:supp_training}\par
\makebox[1.5em][l]{}\ref{subsec:supp_visual_homogeneity_prompt_examples}. Prompt Examples \dotfill \pageref{subsec:supp_visual_homogeneity_prompt_examples}\par
\makebox[1.5em][l]{}\ref{subsec:supp_additional_concept_types}. Additional Concept Types \dotfill \pageref{subsec:supp_additional_concept_types}\par
\makebox[1.5em][l]{}\ref{subsec:supp_additional_vlm_evaluators}. Additional VLM Evaluator \dotfill \pageref{subsec:supp_additional_vlm_evaluators}\par
\makebox[1.5em][l]{}\ref{subsec:supp_diversity_trend_significance}. Statistical Support for Diversity Trends \dotfill \pageref{subsec:supp_diversity_trend_significance}\par
\makebox[1.5em][l]{}\ref{subsec:supp_class_word_priors}. Effect of Class-Word Priors \dotfill \pageref{subsec:supp_class_word_priors}\par

\textbf{\ref{sec:supp_ablation_other_layers}. Dominant-Token Detection in Other Layers} \dotfill \pageref{sec:supp_ablation_other_layers}\par

\textbf{\ref{subsec:supp_temporal_metric_design}. Metric Design for Temporal Analysis} \dotfill \pageref{subsec:supp_temporal_metric_design}\par

\textbf{\ref{sec:supp_attention_intervention}. Preliminary Attention Intervention} \dotfill \pageref{sec:supp_attention_intervention}\par

\textbf{\ref{sec:supp_multi_head_ablation}. Multi-Head Ablation: Across Early Timesteps} \dotfill \pageref{sec:supp_multi_head_ablation}\par
\makebox[1.5em][l]{}\ref{subsec:supp_early_timestep_ablation}. Early-Timestep Ablation \dotfill \pageref{subsec:supp_early_timestep_ablation}\par
\makebox[1.5em][l]{}\ref{subsec:supp_head_ablation_robustness}. Head-Ablation Robustness \dotfill \pageref{subsec:supp_head_ablation_robustness}\par

\textbf{\ref{sec:supp_detection}. DvD Phenomenon Detection} \dotfill \pageref{sec:supp_detection}\par
\makebox[1.5em][l]{}\ref{subsec:supp_detection_methodology}. Detection Methodology \dotfill \pageref{subsec:supp_detection_methodology}\par
\makebox[1.5em][l]{}\ref{subsec:supp_detection_selected_config}. Selected Detection Configuration \dotfill \pageref{subsec:supp_detection_selected_config}\par
\makebox[1.5em][l]{}\ref{subsec:supp_detection_validation}. Validation of Detection Accuracy \dotfill \pageref{subsec:supp_detection_validation}\par
\makebox[1.5em][l]{}\ref{subsec:supp_detection_limitations_future}. Limitations and Future Work \dotfill \pageref{subsec:supp_detection_limitations_future}\par
}%

\clearpage

\section{DominanceBench}
\label{sec:supp_dominancebench}

\subsection{Training-Image Visual Homogeneity Analysis}
\label{subsec:supp_training_data_diversity}

DominanceBench (Sec.~\ref{subsec:db}) categorizes artists, landmarks, and characters as high-homogeneity concepts and common objects as low-homogeneity concepts based on the training-image visual homogeneity hypothesis. Here, we provide empirical support for this categorization by analyzing the relative visual homogeneity of LAION training images for each concept in Table~\ref{tab:vocab}.

\begin{table}[h]
    \centering
    \scriptsize
    \vspace{-2.0em}
    \caption{Concept sets used to collect DominanceBench prompts.}
    \vspace{-1.0em}
    \begin{tabular}{@{}l p{0.88\textwidth}@{}}
      \toprule
      \textbf{Category} & \textbf{Terms} \\
      \midrule
      Artist &
      Amedeo Modigliani, Andy Warhol, Camille Pissarro, Caravaggio, Claude Monet, Edgar Degas, Edvard Munch, Egon Schiele, Frida Kahlo, Gustav Klimt, Hokusai, Kandinsky, Leonardo da Vinci, Leonid Afremov, Monet, Pierre-Auguste Renoir, Raphael, Rembrandt, Titian, Van Gogh, Vermeer \\
      \midrule
      Character &
      Barbie, Batman, Black Panther, Buzz Lightyear, Captain America, Cinderella, Goku, Hello Kitty, Hulk, Iron Man, Joker, Luffy, Mickey Mouse, Minions, Naruto, Pikachu, Rapunzel, Simpson, Spider-Man, Superman, Thor, Wonder Woman, Woody \\
      \midrule
      Landmark &
      Acropolis, Big Ben, Brandenburg Gate, Colosseum, Eiffel Tower, Empire State Building, Golden Gate Bridge, Grand Canyon, Great Wall of China, Hagia Sophia, Kremlin, Machu Picchu, Neuschwanstein Castle, Parthenon, Petra, Pyramid, Space Needle, Taj Mahal, Times Square, Tower Bridge \\
      \midrule
      Object &
      Apron, Bag, Battery Charger, Bottle, Canvas Bag, Carry-All Pouch, Clock, Coaster, Coffee Mug, Cup, Curtain, Halloween, Headphones, Hoodie, Lamp, Magnet, Magnets, Mouse Pad, Mousepad, Mug, Necklace, Notebook, Placemat, Pouch, Rug, Sleeves, Spiral Notebook, Sweatshirt, T-Shirt, Tattoo, Tie, Tote Bag, Umbrella, Wallet \\
      \bottomrule
      \vspace{-2.5em}
    \end{tabular}
    \label{tab:vocab}
\end{table}

For each concept, we keep the top 1,000 most frequent LAION captions containing the keyword and download the corresponding images. We then compute CLIP ViT-L/14 image embeddings and measure intra-category cosine distances to assess relative visual homogeneity.

\begin{figure*}[t]
    \centering
    \begin{subfigure}[b]{\linewidth}
      \includegraphics[width=\linewidth]{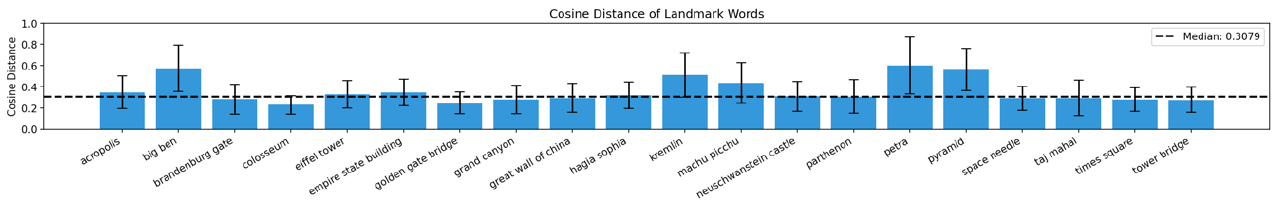}
      \vspace{-1.8em}
      \caption{Landmark (median: 0.3079)}
      \vspace{0.6em}
    \end{subfigure}
    
    \begin{subfigure}[b]{\linewidth}
      \includegraphics[width=\linewidth]{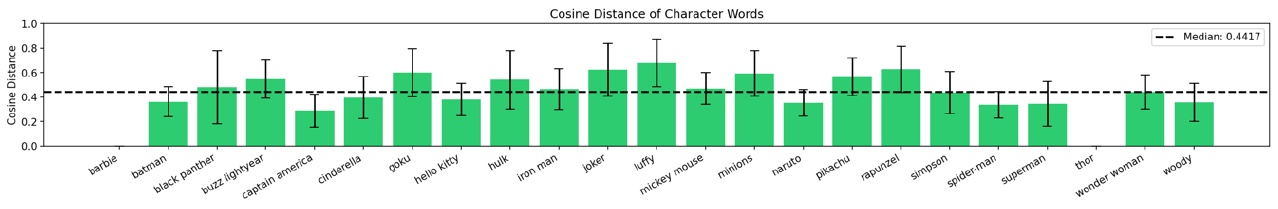}
      \vspace{-1.8em}
      \caption{Character (median: 0.4417)}
      \vspace{0.6em}
    \end{subfigure}
    
    \begin{subfigure}[b]{\linewidth}
      \includegraphics[width=\linewidth]{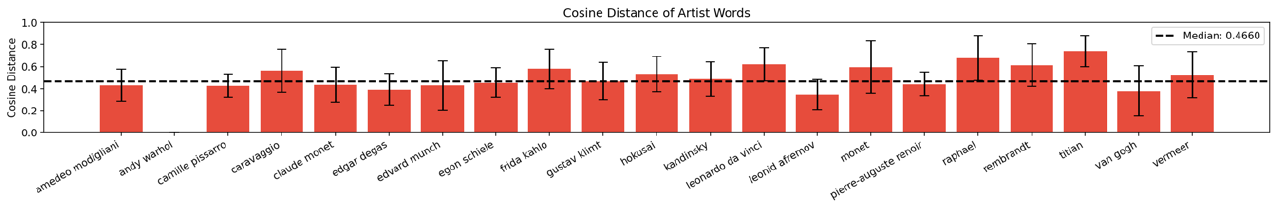}
      \vspace{-1.8em}
      \caption{Artist (median: 0.4660)}
      \vspace{0.6em}
    \end{subfigure}
    
    \begin{subfigure}[b]{\linewidth}
      \includegraphics[width=\linewidth]{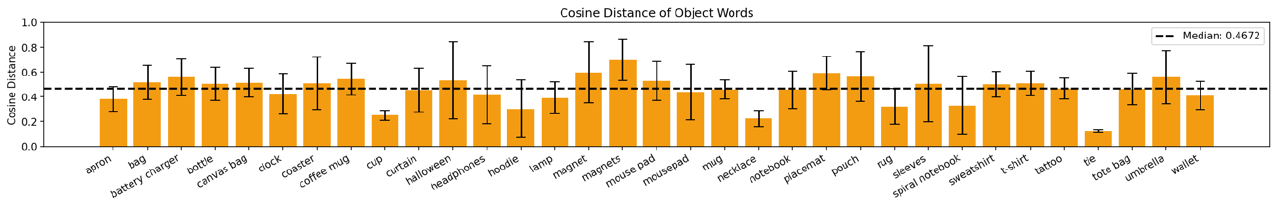}
      \vspace{-1.8em}
      \caption{Object (median: 0.4672)}
      \vspace{0.6em}
    \end{subfigure}
    
    \vspace{-1.0em}
    \caption{Intra-category cosine distance distributions for LAION training images. For each concept in Table~\ref{tab:vocab}, we collect the training images associated with the 1,000 most frequent LAION captions containing the keyword and compute cosine distances in CLIP space. Lower distances indicate higher visual homogeneity.}
    \vspace{-0.7em}
    \label{fig:cos_sim}
\end{figure*}

\cref{fig:cos_sim} shows the cosine distance distribution for each category. We report medians to reduce the influence of near-duplicate samples. Landmark images exhibit the lowest median distance (0.3079), indicating the highest visual homogeneity. Character images follow (median 0.4417), while artist images (median 0.4660) and object images (median 0.4672) display substantially higher intra-category distances, indicating lower visual homogeneity.

Overall, this analysis provides evidence of a relative ordering in training-image visual homogeneity, with landmark images showing the highest homogeneity and object images the lowest.

\subsection{Prompt Collection and Evaluation Details}
\label{subsec:supp_prompt_collection_eval}
DominanceBench prompts rely on four concept sets: $\mathcal{A}$ (artist), $\mathcal{C}$ (character), $\mathcal{L}$ (landmark), and $\mathcal{O}$ (object). Table~\ref{tab:vocab} enumerates their elements, and we denote their union by $\mathcal{V} = \mathcal{A} \cup \mathcal{C} \cup \mathcal{L} \cup \mathcal{O}$. We use these sets to retrieve candidate prompts from LAION and filter them using the DvD-Score-based procedure in \cref{alg:dominancebench_filtering}.

To compute DvD Scores, we use category-specific questions for each concept type. The five questions per category are listed in Table~\ref{tab:questions}.

\begin{table}[!htbp]
  \centering
  \scriptsize
  \caption{Questions used in collecting DominanceBench for each category.}
  \vspace{-1.0em}
  \label{tab:questions}
  \begin{tabular}{@{}l p{0.9\textwidth}@{}}
    \toprule
    \textbf{Category} & \textbf{Questions} \\
    \midrule
    \multirow{5}{*}{Artist} &
    \begin{itemize}[leftmargin=*, topsep=0pt, partopsep=0pt, parsep=0pt, itemsep=2pt]
      \vspace{-0.5\baselineskip}
      \item Is this image painted in the artistic style of \texttt{Artist}, using realistic lighting and smooth brush strokes?
      \item Does the person or subject look painted with the realistic portrait technique typical of \texttt{Artist}?
      \item Does this image show brushwork, color tones, or shading similar to paintings by \texttt{Artist}?
      \item Does this look like a painting created by \texttt{Artist}, rather than a photo or modern artwork?
      \item Is the visual composition, lighting, and realism clearly similar to \texttt{Artist}'s paintings?
    \end{itemize} \\
    \midrule
    \multirow{5}{*}{Landmark} &
      \begin{itemize}[leftmargin=*, topsep=0pt, partopsep=0pt, parsep=0pt, itemsep=2pt]
        \vspace{-0.5\baselineskip}
        \item Does the image show architectural details or a skyline resembling the \texttt{Landmark}?
        \item Is the \texttt{Landmark} the main subject or part of the background in this image?
        \item Is the \texttt{Landmark} or part of it visible anywhere in the image?
        \item Is this a photograph or realistic depiction of the \texttt{Landmark}?
        \item Is the structure clearly identifiable as the \texttt{Landmark}?
      \end{itemize} \\
    \midrule
    \multirow{5}{*}{Character} &
      \begin{itemize}[leftmargin=*, topsep=0pt, partopsep=0pt, parsep=0pt, itemsep=2pt]
        \vspace{-0.5\baselineskip}
        \item Does the image contain a logo, costume, or any reference to \texttt{Character}?
        \item Is the scene related to, inspired by, or set in the world of \texttt{Character}?
        \item Is there a visual element or theme in the image connected to \texttt{Character}?
        \item Does the person or figure look like \texttt{Character}?
        \item Is \texttt{Character} visible in the image?
      \end{itemize} \\
    \midrule
    \multirow{5}{*}{Object} &
      \begin{itemize}[leftmargin=*, topsep=0pt, partopsep=0pt, parsep=0pt, itemsep=2pt]
        \vspace{-0.5\baselineskip}
        \item Does this image look like a product photo or include a depiction of a \texttt{Object}?
        \item Is there a \texttt{Object} included or partly shown in the image?
        \item Is the appearance or shape recognizable as a \texttt{Object}?
        \item Does the image feature or focus on a \texttt{Object}?
        \item Is a \texttt{Object} visible in the image?
      \end{itemize} \\
    \bottomrule
  \end{tabular}
\end{table}

\subsection{Independent Prompt Set}
\label{subsec:supp_independent_prompt_set}
To check whether the DvD behavior in DominanceBench is specific to SD~1.4-based prompt filtering, we construct a 300-prompt independent set from the same concept/object vocabulary in Table~\ref{tab:vocab}, without applying the DvD-Score-based filtering criterion in \cref{alg:dominancebench_filtering}.
The independent prompt set will be released upon publication.

For each prompt, we generate 10 images with different random seeds and compute the per-prompt mean DvD Score.
The fraction of prompts with mean DvD Score at least 36 remains high in this independent set: 92.67\% for SD~1.4 and 50.67\% for SD~2.1, compared with 100.00\% and 70.00\% on DominanceBench, respectively.
This result suggests that the phenomenon is not solely an artifact of SD~1.4-based benchmark construction.

\subsection{Threshold Robustness}
\label{subsec:supp_threshold_robustness}
In Sec.~\ref{subsec:db}, we set the DvD threshold to 36 using the upper quartile of the balanced-prompt score distribution.
To check whether DominanceBench is sensitive to this threshold choice, we vary the threshold while keeping the same 10-seed evaluation protocol.
All 300 DominanceBench prompts would be retained when the threshold varies from 32 to 40; stricter thresholds of 48 and 60 retain 294 and 266 prompts, respectively.
This indicates that the benchmark construction is not sensitive to the exact threshold value of 36.


\section{Reference Prompts}
\label{sec:supp_reference_prompts}

\subsection{Balanced Prompts}
\label{subsec:supp_balanced_prompts}
In Sec.~\ref{subsec:db} and Sec.~\ref{subsubsec:focus_score}, we use balanced prompts to distinguish DvD behavior from successful multi-concept generation.

We leverage the non-memorized benchmark of Ren et al.~\cite{ren2024unveiling}, a reference set used in their memorization comparison, which contains 500 prompts describing common subjects and objects.
For each prompt, we ask GPT-5~\citeS{singh2025openai} to identify the major subjects or objects that should be visually represented in the generated image. We retain only prompts for which it identifies exactly two concepts, yielding 300 prompts.
This filtered subset matches the two-concept structure of DominanceBench.

Because the balanced prompt set contains diverse common concepts, we evaluate it using the generic questions in Table~\ref{tab:balanced_vqa_questions} about whether each concept is present, rather than the category-specific templates in Table~\ref{tab:questions}.
We therefore use its DvD Score distribution primarily as a reference distribution for threshold calibration, as shown in Fig.~\ref{fig:dvd_comparison}, rather than as a strictly matched absolute benchmark against DominanceBench.

\begin{table}[t]
    \centering
    \caption{Questions used in calculating DvD Score for balanced prompts.}
    \vspace{-1.0em}
    \label{tab:balanced_vqa_questions}
    \resizebox{0.8\linewidth}{!}{%
      \begin{tabular}{@{}p{0.18\linewidth} p{0.78\linewidth}@{}}
        \toprule
        \textbf{Category} & \textbf{Questions} \\
        \midrule
        \multirow{5}{*}{Concept} &
          \begin{itemize}[leftmargin=*, topsep=0pt, partopsep=0pt, parsep=0pt, itemsep=2pt]
            \vspace{-0.5\baselineskip}
            \item Does the image include \texttt{Concept} as part of the scene?
            \item Does the image show \texttt{Concept} in a recognizable way?
            \item Is \texttt{Concept} visually represented in the image?
            \item Can you see \texttt{Concept} clearly in the image?
            \item Is \texttt{Concept} present in the image?
          \end{itemize} \\
        \bottomrule
      \end{tabular}%
    }
  \end{table}

\subsection{Memorized Prompts}
\label{subsec:supp_memorized_prompts}
In Sec.~\ref{subsec:head_ablation}, we compare DvD with memorization using 500 memorized prompts identified by Carlini et al.~\cite{carlini2023extracting}. Among existing studies on memorization, Wen et al.~\cite{wen2024detectingexplainingmitigatingmemorization} proposed a detection method that computes the L2 norm of text-conditional noise predictions at the first denoising step. A higher L2 norm value indicates that the prompt consistently reproduces the same image regardless of the random seed. To further characterize DominanceBench, we apply this metric to three prompt sets: memorized prompts, DominanceBench, and balanced prompts.

\begin{figure}[t]
    \centering
    \includegraphics[width=0.7\textwidth]{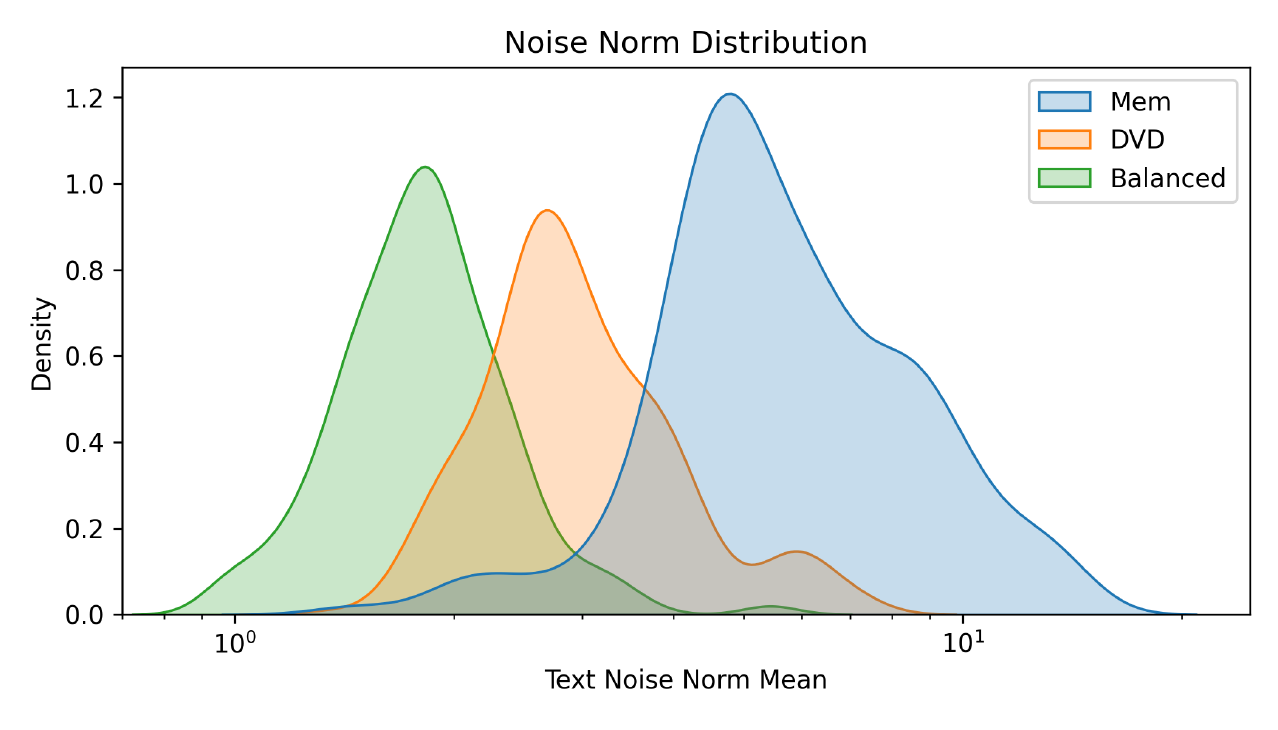}
    \vspace{-1.2em}
    \caption{L2 norm distributions of text-conditional noise predictions at the first denoising step across memorized, DominanceBench, and balanced prompts.}
    \vspace{0.0mm}
    \label{fig:l2norm}
\end{figure}

Fig.~\ref{fig:l2norm} shows the L2 norm distributions for these three prompt sets. Memorized prompts show the highest values, DominanceBench shows intermediate values, and balanced prompts show the lowest. This intermediate position suggests that DominanceBench shares some generation-time characteristics with memorization.

\section{Visual Homogeneity Experiments: Details and Extensions}
\label{sec:supp_training}

\subsection{Prompt Examples}
\label{subsec:supp_visual_homogeneity_prompt_examples}
In \cref{subsec:toy_example}, we evaluate the learned concept (e.g., ``sks castle'') together with a partner concept in two-concept prompts. We include both a \textit{normal} order and a \textit{flipped} order for each concept pair, and compute DvD Scores over 10 random seeds.
Table~\ref{tab:supp_prompt_examples} lists representative prompt examples used for evaluation.

\begin{table*}[t]
    \centering
    \caption{Representative prompt examples used in the visual homogeneity experiments.}
    \vspace{-0.7em}
    \label{tab:supp_prompt_examples}
    \begin{tabular}{@{}lll p{0.58\textwidth}@{}}
      \toprule
      \textbf{Class} & \textbf{Mode} & \textbf{Partner Concept} & \textbf{Prompt} \\
      \midrule
      castle & normal & fountain & A sks castle with a fountain nearby \\
       & flipped & fountain & A fountain next to a sks castle \\
      mountain & normal & flag & A sks mountain with a flag nearby \\
       & flipped & flag & A flag next to a sks mountain \\
      arena & normal & statue & A sks arena with a statue nearby \\
       & flipped & statue & A statue next to a sks arena \\
      minions & normal & street lamp & A sks minions with a street lamp nearby \\
       & flipped & street lamp & A street lamp next to a sks minions \\
      van gogh & normal & bench & A bench in the style of sks van gogh \\
       & flipped & bench & In the style of sks van gogh, a bench \\
      \bottomrule
    \end{tabular}
\end{table*}

\subsection{Additional Concept Types}
\label{subsec:supp_additional_concept_types}
To examine whether the trend observed in \cref{subsec:toy_example} extends beyond the landmark-based concepts, we repeat the same controlled homogeneity experiment on two additional concept types: the character concept \textit{Minions} and the artist-style concept \textit{Van Gogh}.

\begin{figure*}[t]
    \centering
    \begin{subfigure}[b]{0.49\linewidth}
        \centering
        \includegraphics[width=\linewidth]{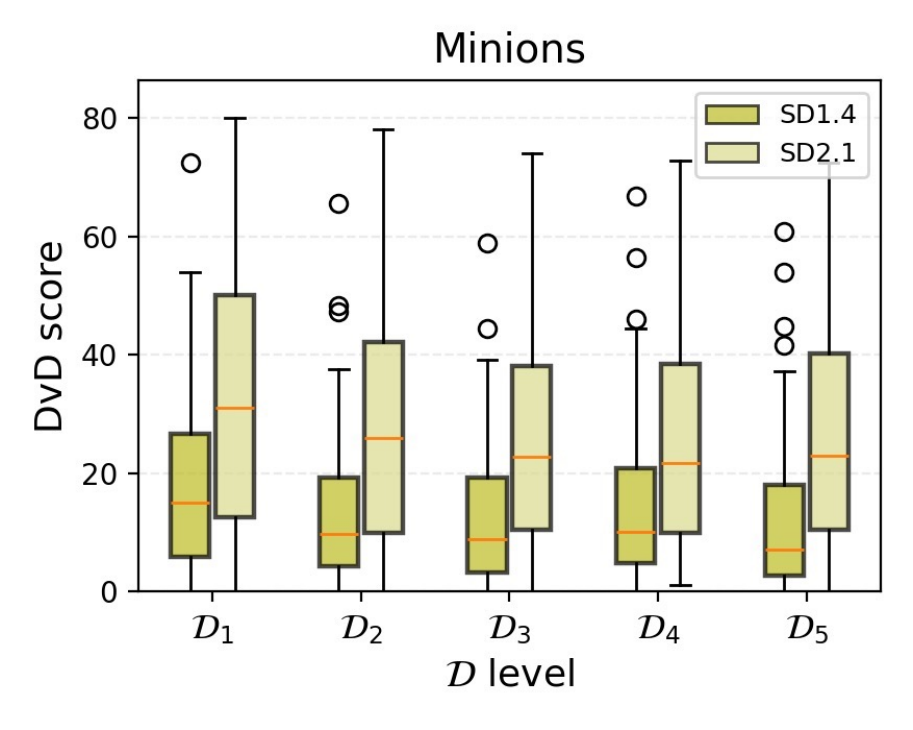}
        \caption{Minions}
    \end{subfigure}
    \hfill
    \begin{subfigure}[b]{0.49\linewidth}
        \centering
        \includegraphics[width=\linewidth]{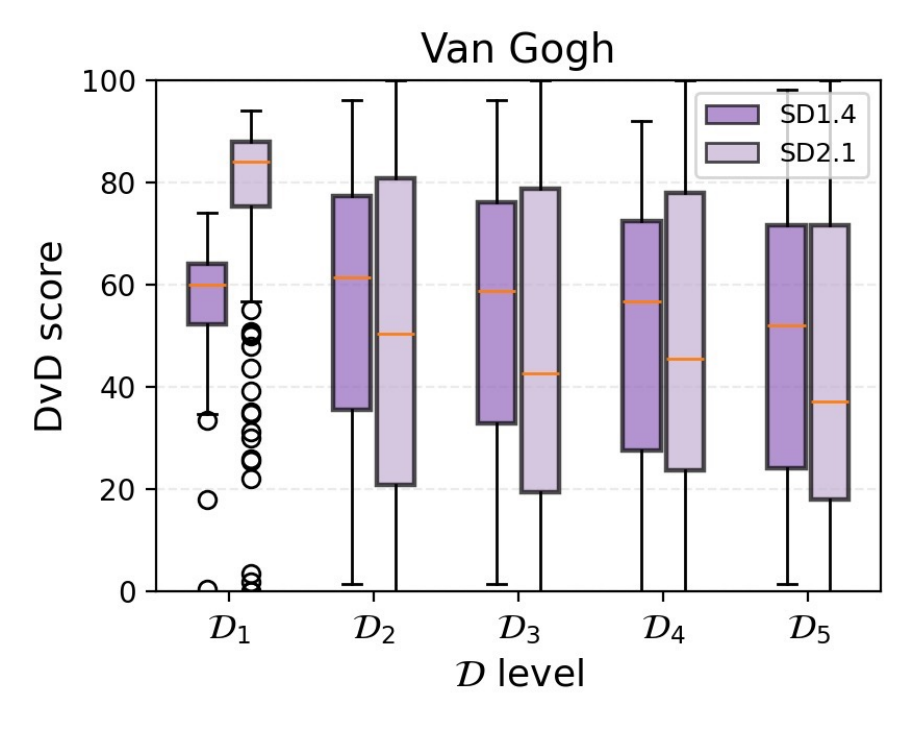}
        \caption{Van Gogh}
    \end{subfigure}
    \vspace{-0.7em}
    \caption{Effect of training-image visual homogeneity on dominance for two additional concept types: the character concept \textit{Minions} and the artist-style concept \textit{Van Gogh}. For both SD~1.4 and SD~2.1, we show the distributions of per-prompt mean DvD Scores across homogeneity configurations. Higher scores in more homogeneous settings indicate stronger dominance.}
    \label{fig:add_concept_types}
\end{figure*}

Following \cref{subsec:toy_example}, for each concept we construct five 120-image training sets, $\mathcal{D}_k$ ($k=1,\ldots,5$), where smaller $k$ denotes a more visually homogeneous set.
For each $\mathcal{D}_k$, we fine-tune only the UNet of SD~1.4 and SD~2.1 while keeping the text encoder fixed.
We also use the same prompt construction, normal/flipped concept orders, and DvD Score computation as in \cref{subsec:toy_example}.
For \textit{Minions}, we evaluate whether the generated image contains both the learned character and the partner concept. For \textit{Van Gogh}, we evaluate whether the generated image preserves the learned style together with the partner concept.
Figure~\ref{fig:add_concept_types} shows the distributions of per-prompt mean DvD Scores across homogeneity configurations for the two additional concepts.
For both SD~1.4 and SD~2.1, the score distributions for \textit{Minions} and \textit{Van Gogh} shift toward higher values as training-image visual homogeneity increases.
This trend is most pronounced in the most homogeneous setting ($\mathcal{D}_1$), where the learned concept more frequently dominates the partner concept.
In contrast, less homogeneous settings yield more balanced generations.

These results suggest that the relationship between visual homogeneity and dominance is not limited to the landmark-based concepts used in the main paper, but also extends to character and artist-style concepts.

\subsection{Additional VLM Evaluator}
\label{subsec:supp_additional_vlm_evaluators}

To test whether the conclusion in \cref{subsec:toy_example} depends on the choice of VLM evaluator, we recompute DvD Scores using InternVL2.5-8B~\citeS{chen2024expanding} in place of Qwen2.5-VL, while leaving the generated images, prompts, homogeneity configurations, and concept-presence questions unchanged.
This experiment tests the robustness of the observed relationship between training-image visual homogeneity and dominance to the choice of VLM evaluator.

\noindent\textbf{Experimental Setup.}
We reuse the generated images from the homogeneity-controlled training sets $\mathcal{D}_k$ ($k=1,\ldots,5$) for the three learned concepts: \textit{castle}, \textit{mountain}, and \textit{arena}.
For each image, we use InternVL2.5-8B to reassess the visual presence of the two concepts and compute the DvD Score using the same procedure as in \cref{subsec:toy_example}.
As in \cref{subsec:toy_example}, we average the scores over 10 random seeds and report the distributions of per-prompt mean DvD Scores separately for SD~1.4 and SD~2.1.

\begin{figure*}[t]
    \centering
    \begin{subfigure}[b]{0.32\linewidth}
        \centering
        \includegraphics[width=\linewidth]{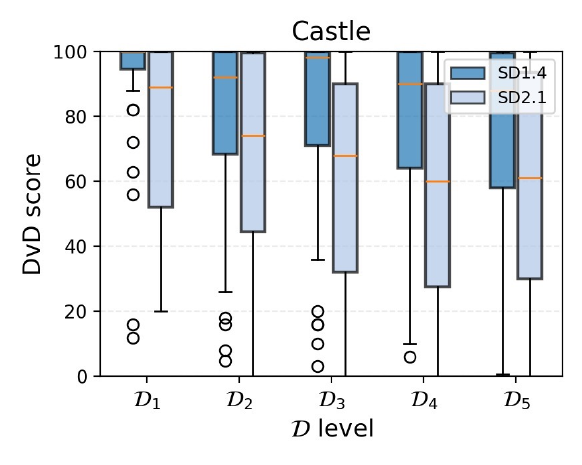}
        \caption{Castle}
    \end{subfigure}
    \hfill
    \begin{subfigure}[b]{0.32\linewidth}
        \centering
        \includegraphics[width=\linewidth]{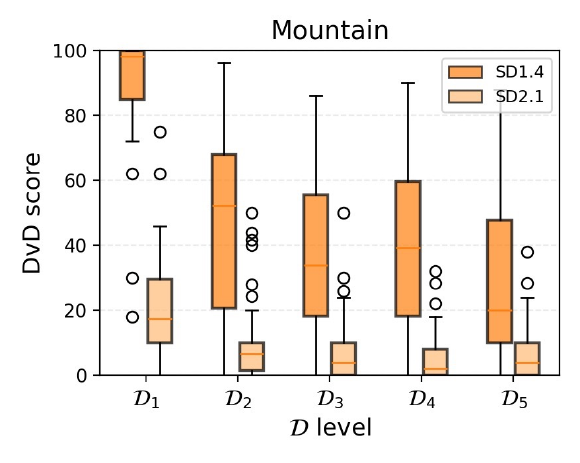}
        \caption{Mountain}
    \end{subfigure}
    \hfill
    \begin{subfigure}[b]{0.32\linewidth}
        \centering
        \includegraphics[width=\linewidth]{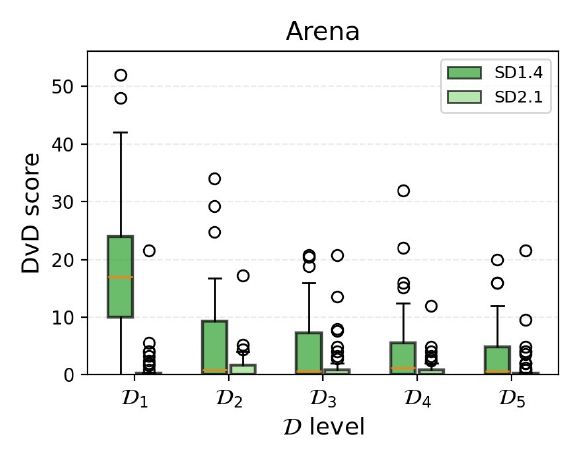}
        \caption{Arena}
    \end{subfigure}
    \vspace{-0.7em}
    \caption{DvD Score distributions recomputed with InternVL2.5-8B as the VLM evaluator for the three learned concepts.}
    \vspace{-0.7em}
    \label{fig:internvl_vlm_eval}
\end{figure*}

\noindent\textbf{Results.}
Figure~\ref{fig:internvl_vlm_eval} shows that the InternVL2.5-8B-based scores follow the same qualitative trend as \cref{fig:toy_example_combined}.
For both SD~1.4 and SD~2.1, the DvD Score distributions tend to shift toward higher values as training-image visual homogeneity increases (smaller $k$ in $\mathcal{D}_k$).
Overall, this evaluation supports the same conclusion that greater visual homogeneity is associated with stronger dominance.

While these extended evaluations support the qualitative relationship between training-image visual homogeneity and dominance, two possible confounds remain.
We therefore provide additional checks on whether the diversity-dependent trend is statistically reliable and whether it can be explained by pretrained class-word priors rather than the diversity manipulation itself.

\subsection{Statistical Support for Diversity Trends}
\label{subsec:supp_diversity_trend_significance}
To move beyond visual inspection of Fig.~\ref{fig:toy_example_combined}, we quantify the diversity-dependent trends using bootstrap 95\% confidence intervals and one-sided Spearman permutation tests across homogeneity configurations.
Table~\ref{tab:diversity_trend_significance} reports the DvD Score drop from the most homogeneous setting $\mathcal{D}_1$ to the most diverse setting $\mathcal{D}_5$, together with the one-sided Spearman $p$-value.

\begin{table}[t]
\centering
\small
\caption{Statistical test results for the visual-homogeneity experiments. Each cell reports the DvD Score drop from $\mathcal{D}_1$ to $\mathcal{D}_5$ / one-sided Spearman permutation-test $p$-value.}
\vspace{-0.8em}
\label{tab:diversity_trend_significance}
\begin{tabular}{lccc}
\toprule
\textbf{Model} & \textbf{Castle} & \textbf{Mountain} & \textbf{Arena} \\
\midrule
SD~1.4 & 17.48 / 0.0005 & 29.93 / 0.0005 & 14.02 / 0.0005 \\
SD~2.1 & 19.23 / 0.0005 & 7.29 / 0.0005 & 0.78 / 0.1599 \\
\bottomrule
\end{tabular}
\end{table}

The trend is significant in 5 of the 6 concept/model panels.
The only exception is Arena under SD~2.1, where the DvD Score range is small.

\subsection{Effect of Class-Word Priors}
\label{subsec:supp_class_word_priors}
A second possible confound is that the learned concepts are expressed using class words such as ``castle,'' ``mountain,'' and ``arena,'' which may already carry pretrained visual priors in the base model.
We therefore check whether these pretrained class-word priors alone explain the diversity-dependent trend in Sec.~\ref{subsec:toy_example}.
Using the same prompts with base SD models before fine-tuning, only 0--18\% of prompts exceed the DvD threshold, depending on the class and model.
Moreover, all homogeneity configurations for a given learned concept share the same class word.
Thus, class-word priors may affect the absolute score level, but they do not by themselves explain the diversity-dependent trend observed across $\mathcal{D}_1$--$\mathcal{D}_5$.


\section{Dominant-Token Detection in Other Layers}
\label{sec:supp_ablation_other_layers}

In \cref{subsubsec:focus_score}, we analyze dominant-token signals in SD~2.1 using layers 8--10. 
This choice follows the layer-wise trend observed in \cref{fig:layerwise_focus}, which shows that the gap in focus scores between DominanceBench and balanced prompts peaks around layers 8--10 in both SD~1.4 and SD~2.1.

However, when viewed in isolation, \cref{fig:focus_sd21} may give a different impression: the apparent gap between the balanced and DvD cases appears larger in layers 5--7 than in layers 8--10. To better understand where dominant-token signals are most consistently captured, we conduct an additional layer-wise analysis.

For each layer (and layer group), we measure how frequently dominant tokens are detected across 300 prompts. \cref{tab:layer_sweep_sd21} summarizes the results. 
When analyzed individually, layer 10 yields the highest detection rate (72.0\%), indicating that dominant-token signals are most clearly detected at this layer. 
Because layers 5 and 6 correspond to the last downblocks, we additionally report their grouped detection rate. When grouping layers, the average detection rate of layers 8--10 (54.0\%) is substantially higher than that of layers 5--6 (26.00\%).

Overall, while the last downblocks (layers 5--6) may show larger differences between the balanced and DvD cases in \cref{fig:focus_sd21}, dominant-token signals are more consistently detected when using layers 8--10.
For this reason, the SD~2.1 analysis in \cref{subsubsec:focus_score} focuses on layers 8--10.

\begin{table*}[t]
\centering
\small
\caption{Layer-wise detection rate of dominant tokens in SD~2.1 across 300 prompts. 
``Included'' denotes the number of prompts in which the dominant concept token 
is identified as the maximally attended token in the corresponding layer.}
\resizebox{0.99\textwidth}{!}{
\begin{tabular}{lccccccccc}
\toprule
Layer Setting & 5 & 6 & 7 & 8 & 9 & 10 & 5--6 & 8--10 & 5--10 \\
\midrule
Included
& 103
& 71
& 30
& 103
& 137
& 216
& 78
& 162 
& 98\\

Percent 
& 34.33\% 
& 23.67\% 
& 10.00\% 
& 34.33\% 
& 45.67\% 
& 72.00\% 
& 26.00\% 
& 54.00\% 
& 32.67\% \\
\bottomrule
\end{tabular}
}
\label{tab:layer_sweep_sd21}
\end{table*}


\section{Metric Design for Temporal Analysis}
\label{subsec:supp_temporal_metric_design}
While the focus score (Eq.~\ref{eq:focus_score}) uses entropy normalization, our temporal analysis (Sec.~\ref{subsubsec:temporal_analysis}) uses only attention deviation without entropy. We explain why below.

\subsection{Different Objectives Require Different Metrics}
\label{subsec:different_objectives_require_different_metrics}
\label{subsubsec:supp_temporal_objectives_metrics}

The focus score compares attention patterns across different prompts with varying characteristics. The numerator measures attention concentration on the peak token relative to others, while the denominator uses entropy to capture the overall dispersion of the attention distribution. Normalizing entropy by $\log_2 N$ accounts for prompt length, enabling fair comparison across prompts with different token counts.

In contrast, Temporal Analysis tracks attention dynamics within a single prompt over time. The attention deviation $\alpha_i^{(\ell,t)} = a_i^{(\ell,t)} - \bar{a}_{others}^{(\ell,t)}$ already captures relative token importance, and its temporal change $\Delta\alpha_i^{(\ell,t)}$ directly measures how the competitive balance shifts between concepts. We do not use entropy normalization because it causes distortion from irrelevant tokens, as explained below.

\subsection{Entropy Causes Noise in Temporal Analysis}
\label{subsec:entropy_causes_noise_in_temporal_analysis}
\label{subsubsec:supp_temporal_entropy_noise}

Entropy $H$ measures the dispersion of the entire attention distribution, including tokens irrelevant to the dominant and dominated concepts (denoted as $C_D$ and $C_d$, respectively). When irrelevant tokens shift attention, entropy changes even if the attention to $C_D$ and $C_d$ remains unchanged, leading to distorted interpretations of concept-level dynamics.

As a concrete example, consider 10 tokens ($C_D$, $C_d$, and 8 other tokens) with $\sum a_i = 1.0$ (Table~\ref{tab:noise_example}):

\begin{table}[t]
    \caption{Attention weights at consecutive timesteps. The attention of $C_D$ and $C_d$ remains constant while that of an irrelevant token $T_1$ suddenly increases.}
    \vspace{-1.0em}
    \label{tab:noise_example}
    \centering
    \resizebox{0.6\linewidth}{!}{%
      \begin{tabular}{@{}lccc@{}}
        \toprule
        Token & Timestep $t$ & Timestep $t+1$ & Change \\
        \midrule
        $C_D$ (Dominant) & \textbf{0.40} & \textbf{0.40} & \textbf{0} \\
        $C_d$ (Dominated) & 0.30 & 0.30 & 0 \\
        $T_1$ (Another token) & 0.05 & \textbf{0.15} & \textcolor{red}{+0.10} \\
        Other 7 tokens & 0.036 each & 0.021 each & $-0.015$ each \\
        \bottomrule
      \end{tabular}%
    }
\end{table}

\noindent\textbf{Using attention deviation (our approach):}
{\small
\begin{gather}
    \bar{a}_{others}^{(t)} = \frac{1-0.40}{9} \approx 0.067 \\
    \bar{a}_{others}^{(t+1)} = \frac{1-0.40}{9} \approx 0.067 \\
    \alpha_{C_D}^{(t)} = 0.40 - 0.067 = 0.333 \\
    \alpha_{C_D}^{(t+1)} = 0.40 - 0.067 = 0.333 \\
    \Delta\alpha_{C_D} = 0
\end{gather}
}

\textit{Correct interpretation:} $C_D$'s relative advantage is unchanged.

\noindent\textbf{If entropy normalization is used hypothetically:}

At timestep $t$, attention is concentrated primarily on $C_D$ (0.40) and $C_d$ (0.30), with other tokens having small weights (e.g., 0.05, 0.036 each), yielding low $H_t$. 
At timestep $t+1$, the attention of $T_1$ increases to 0.15, while those of $C_D$ (0.40) and $C_d$ (0.30) remain constant, yielding higher $H_{t+1}$.

An entropy-normalized metric would give:
{\small
\begin{gather}
    \text{Score}_t = \frac{\alpha_{C_D}^{(t)}}{H_t / \log_2 N} \quad \text{(high value)} \\
    \text{Score}_{t+1} = \frac{\alpha_{C_D}^{(t+1)}}{H_{t+1} / \log_2 N} \quad \text{(lower value)} \\
    \Delta\text{Score} < 0
\end{gather}
}

\textit{Distorted interpretation:} ``$C_D$'s dominance decreased''---despite $a_{C_D}$ remaining constant at 0.40. This distortion occurs because the entropy increase stems from irrelevant token $T_1$ (0.05 $\rightarrow$ 0.15), not from changes in the dominant ($C_D$) or dominated ($C_d$) concepts that Temporal Analysis aims to isolate.

\subsection{Additional Layer-Wise Consideration}
\label{subsec:additional_layer_wise_consideration}
\label{subsubsec:supp_temporal_layerwise_considerations}

Beyond the token-level noise discussed above, incorporating entropy would create an additional issue for cross-layer tracking. As described in Sec.~\ref{subsubsec:temporal_analysis}, we track dominant concept tokens in layers 8--10 and dominated tokens in layer 7. Since each layer has its own entropy value, any metric involving entropy cannot be directly compared across layers. Attention deviation $\alpha_i^{(l,t)} = a_i^{(l,t)} - \bar{a}_{others}^{(l,t)}$ isolates relative token importance within each layer, enabling consistent cross-layer comparison.

\noindent\textbf{Summary.} By using only attention deviation, our Temporal Analysis (1) maintains direct interpretability ($\Delta\alpha$ = change in relative advantage), (2) eliminates noise from irrelevant tokens, and (3) enables consistent cross-layer tracking of the dominant and dominated concepts that characterize the DvD phenomenon.


\section{Preliminary Attention Intervention}
\label{sec:supp_attention_intervention}

Motivated by the early-step attention imbalance observed in Sec.~\ref{subsubsec:temporal_analysis}, we conduct a small perturbation study to probe its functional connection to DvD outcomes.
Specifically, we rescale the cross-attention weights of selected dominant and dominated concept tokens during early denoising steps.

Across 300 DominanceBench prompts and 3 seeds on SD~1.4, the mean DvD Score decreases from 80.08 to 77.29 when boosting dominated-token attention, to 78.73 when suppressing dominant-token attention, and to 74.76 when applying both changes.
These partial reductions provide preliminary evidence that early attention imbalance is functionally related to DvD outcomes.


\section{Multi-Head Ablation: Across Early Timesteps}
\label{sec:supp_multi_head_ablation}

\subsection{Early-Timestep Ablation}
\label{subsec:supp_early_timestep_ablation}

In \cref{subsec:head_ablation}, we ablate specific heads throughout the full denoising process to examine whether each phenomenon is localized to a small set of heads or distributed across many heads.
However, \cref{subsubsec:temporal_analysis} shows that the attention imbalance underlying DvD is already established within the first few denoising steps.
Here, we check whether the same head-level pattern remains when the analysis is restricted to the early denoising steps only.

We repeat the multi-head ablation protocol in \cref{subsec:head_ablation} over early timesteps. The only change is that ablation is applied from timesteps 50 to 40. As in \cref{subsec:head_ablation}, we evaluate single-, dual-, and triple-head ablations in layers 1--6 and classify the resulting outputs as \textit{Mitigated}, \textit{Unchanged}, or \textit{Others}.

\begin{figure*}[t]
    \centering
    \begin{subfigure}[b]{0.32\linewidth}
        \centering
        \includegraphics[width=\linewidth]{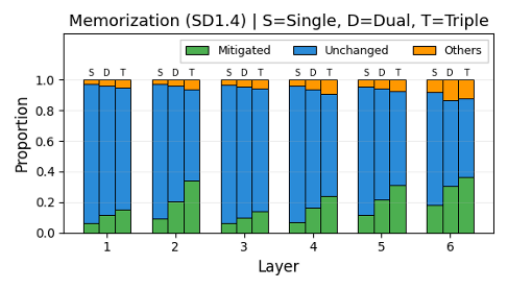}
        \caption{Memorization (SD 1.4)}
    \end{subfigure}
    \hfill
    \begin{subfigure}[b]{0.32\linewidth}
        \centering
        \includegraphics[width=\linewidth]{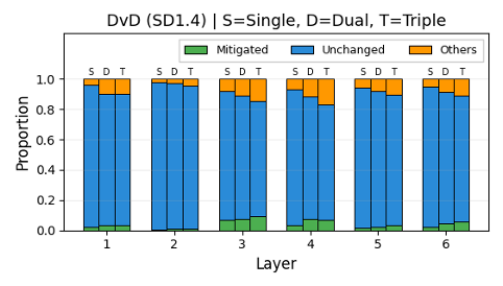}
        \caption{DvD (SD 1.4)}
    \end{subfigure}
    \hfill
    \begin{subfigure}[b]{0.32\linewidth}
        \centering
        \includegraphics[width=\linewidth]{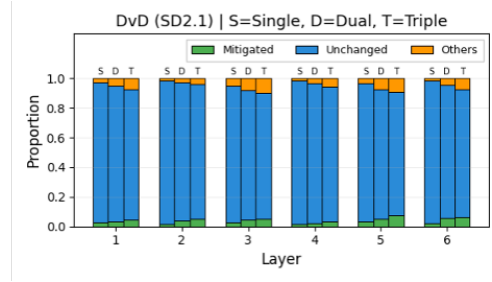}
        \caption{DvD (SD 2.1)}
    \end{subfigure}
    \vspace{-0.7em}
    \caption{Multi-head ablation outcome proportions across layers 1--6 under early-timestep ablation (timesteps 50 to 40).}
    \label{fig:early_timestep_head_ablation}
\end{figure*}

\cref{fig:early_timestep_head_ablation} shows that the qualitative contrast between memorization and DvD remains similar in this early-timestep setting.
For memorization, increasing the ablation size mainly shifts outcomes from \textit{Unchanged} to \textit{Mitigated}, which is consistent with a more localized head-level mechanism.
In contrast, for DvD on SD~1.4, increasing the ablation size yields only a marginal increase in \textit{Mitigated}; the more noticeable change is a shift from \textit{Unchanged} to \textit{Others}.
DvD on SD~2.1 shows the same overall tendency: \textit{Unchanged} remains dominant across most layers and ablation sizes, while \textit{Mitigated} cases remain rare.

Overall, these results are consistent with the view that, even in the early denoising steps, DvD is supported by distributed contributions across heads rather than by a small set of critical heads.

\subsection{Head-Ablation Robustness}
\label{subsec:supp_head_ablation_robustness}
We additionally check whether the head-ablation results in Sec.~\ref{subsec:head_ablation} are sensitive to the ablation strength or to the thresholds used for outcome classification.
For ablation strength, we use 20 SD~1.4 DominanceBench cases from the effective single-head ablation layers (layers 3 and 4) and sweep $\varepsilon$ from $10^{-5}$ to $10^{-1}$.
The mitigated ratio remains stable across this range (28.1--28.8\%).

We also reclassify the existing head-ablation outputs over a threshold grid using LPIPS thresholds 0.5/0.6/0.7, SSCD thresholds 0.4/0.5/0.6, and Qwen concept-presence thresholds 3/4.
Across these 90 settings, the layer with the strongest effect remains within early layers 1--6 in 78 settings.
This suggests that the layer-level trend is not tied to a single ablation strength or classification-threshold choice.


\section{DvD Phenomenon Detection}
\label{sec:supp_detection}

This section explores whether our analysis results can be used for DvD detection during generation. In particular, we show that the focus score can serve as an early-step diagnostic for detection. While the main contribution of this paper is identifying the causes and mechanisms of the DvD phenomenon, from a practical perspective, early detection of DvD during generation can serve as a foundation for future mitigation research.

\subsection{Detection Methodology}
\label{subsec:supp_detection_methodology}

We use the focus score (Eq.~\ref{eq:focus_score}) to detect DvD. When the focus score exceeds a specific threshold, we determine that the prompt is likely to trigger DvD. Specifically, we compute the focus score in low-resolution layers at the first denoising step ($t=50$), and treat the token receiving maximum attention as the detected dominant concept token.

\subsection{Selected Detection Configuration}
\label{subsec:supp_detection_selected_config}

Based on Sec.~\ref{subsubsec:focus_score} (\cref{fig:layerwise_focus}) showing that DominanceBench prompts exhibit particularly high focus scores in layers 8--10, we consider these low-resolution layers and tune the detection threshold. For SD~1.4 and SD~2.1, we sweep the threshold from 0.002 to 0.006 (step 0.001) and measure detection rates (\% of prompts identified as DvD) on DominanceBench ($N_{\text{DvD}}=300$) and balanced prompts ($N_{\text{balanced}}=300$).

Table~\ref{tab:detection_rates_sd14_sd21} reports detection rates under several layer configurations in layers 8--10.
Using layers 9 and 10 yields the best detection performance, measured by the maximal difference between correct detection for DominanceBench prompts and false detection for balanced prompts (40.33\%p with threshold 0.006 for SD 1.4 and 32.00\%p with threshold 0.002 for SD 2.1).

\begin{table*}[t]
  \centering
  \caption{Detection rates (\% of prompts identified as DvD) across focus score thresholds and layer configurations in layers 8--10 (DominanceBench: $N_{\text{DvD}}=300$; balanced: $N_{\text{balanced}}=300$).}
  \vspace{-1.0em}
  \label{tab:detection_rates_sd14_sd21}
  \begin{subtable}[t]{0.49\textwidth}
    \centering
    \caption{SD~1.4}
    \vspace{-0.5em}
    {\setlength{\tabcolsep}{2pt}%
    \renewcommand{\arraystretch}{1.0}%
    \fontsize{5.5pt}{6.5pt}\selectfont
    \begin{tabular}{@{}lccccccc@{}}
      \toprule
      thr & max\_mean & L8 & L9 & L10 & 8\&9 & 8\&10 & 9\&10 \\
      \midrule
      \multicolumn{8}{c}{\textit{DominanceBench (\%)}} \\
      \midrule
      0.004 & 88.33 & 71.00 & 71.33 & 72.67 & 74.00 & 75.00 & 79.33 \\
      0.005 & 82.67 & 56.67 & 61.33 & 67.00 & 63.00 & 70.00 & 71.33 \\
      0.006 & 78.33 & 50.33 & 49.33 & 59.67 & 50.67 & 60.00 & \textbf{65.33} \\
      \midrule
      \multicolumn{8}{c}{\textit{Balanced (\%)}} \\
      \midrule
      0.004 & 70.33 & 60.00 & 48.33 & 29.67 & 57.00 & 50.67 & 42.33 \\
      0.005 & 60.33 & 48.67 & 40.00 & 21.33 & 46.67 & 34.67 & 31.67 \\
      0.006 & 49.33 & 38.00 & 32.00 & 15.00 & 35.67 & 23.67 & \textbf{25.00} \\
      \bottomrule
    \end{tabular}}
  \end{subtable}
  \hfill
  \begin{subtable}[t]{0.49\textwidth}
    \centering
    \caption{SD~2.1}
    \vspace{-0.5em}
    {\setlength{\tabcolsep}{2pt}%
    \renewcommand{\arraystretch}{1.0}%
    \fontsize{5.5pt}{6.5pt}\selectfont
    \begin{tabular}{@{}lccccccc@{}}
      \toprule
      thr & max\_mean & L8 & L9 & L10 & 8\&9 & 8\&10 & 9\&10 \\
      \midrule
      \multicolumn{8}{c}{\textit{DominanceBench (\%)}} \\
      \midrule
      0.002 & 92.33 & 79.00 & 68.33 & 55.00 & 77.67 & 70.67 & \textbf{67.00} \\
      0.003 & 65.67 & 33.33 & 46.33 & 37.33 & 41.67 & 37.00 & 46.00 \\
      0.004 & 45.00 & 15.33 & 32.67 & 26.00 & 24.33 & 21.33 & 28.00 \\
      \midrule
      \multicolumn{8}{c}{\textit{Balanced (\%)}} \\
      \midrule
      0.002 & 99.00 & 98.67 & 36.67 & 36.33 & 90.67 & 88.00 & \textbf{35.00} \\
      0.003 & 80.33 & 70.33 & 15.33 & 21.67 & 34.00 & 32.67 & 19.33 \\
      0.004 & 43.00 & 30.33 & 8.67 & 13.33 & 9.00 & 8.67 & 9.33 \\
      \bottomrule
    \end{tabular}}
  \end{subtable}
\end{table*}

\subsection{Validation of Detection Accuracy}
\label{subsec:supp_detection_validation}

To verify whether our detection method accurately identifies dominant concept tokens, we observe changes in DvD Score after modifying the detected tokens. Specifically, we apply the chosen thresholds (SD~1.4: 0.006; SD~2.1: 0.002) to DominanceBench prompts, identify the token receiving maximum attention in layers 9\&10, and modify it.

Based on the training-image visual homogeneity hypothesis revealed in Sec.~\ref{subsec:toy_example}, we replace detected dominant tokens with generic category names such as:
\begin{itemize}[leftmargin=*,topsep=3pt,itemsep=2pt]
    \item Artist: Van Gogh $\rightarrow$ ``artist''
    \item Landmark: Colosseum $\rightarrow$ ``landmark''
    \item Character: Spider-Man $\rightarrow$ ``character''
\end{itemize}
If detection is accurate, this modification should reduce the DvD Score by replacing specific, high-homogeneity concepts with generic, low-homogeneity terms that allow more flexible representations.

\begin{figure}[t]
  \centering
  \includegraphics[width=0.5\linewidth]{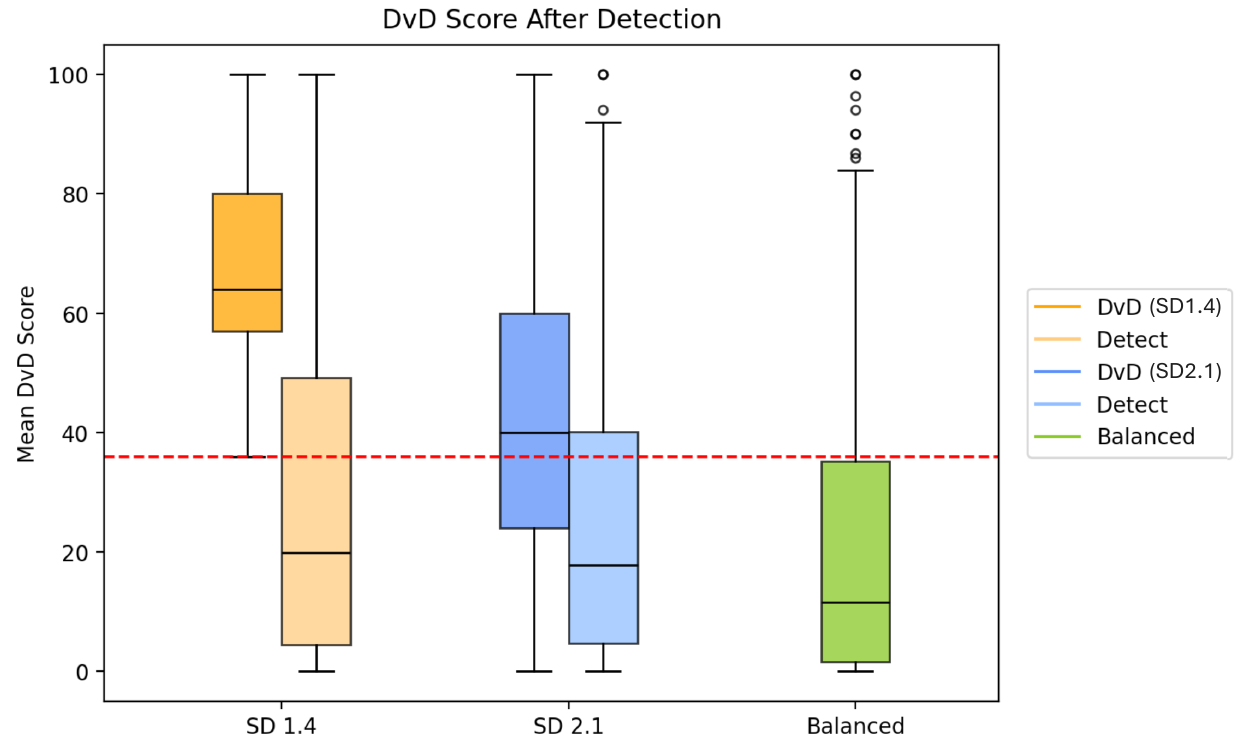}
  \vspace{-0.7em}
  \caption{DvD Score distribution before (darker boxes) and after (lighter boxes) replacing detected tokens with generic category names. Results are shown for DominanceBench prompts on SD~1.4 (orange) and SD~2.1 (blue), with balanced prompts (green) for reference. The observed score reduction suggests that the detected token actually contributes to dominance. (Red dashed line: DvD threshold of 36.)}
  \label{fig:mitigation}
\end{figure}

\cref{fig:mitigation} shows that replacing the detected dominant token substantially reduces DvD Scores on both SD~1.4 and SD~2.1, supporting accurate detection.

\subsection{Limitations and Future Work}
\label{subsec:supp_detection_limitations_future}

The purpose of this section was to validate the accuracy of the detection method, and we clarify that prompt modification itself may not be a practical solution. When a user requests ``Van Gogh coaster,'' generating ``artist coaster'' undermines the original intent. Future research should explore architectural modifications or training-level methods that alleviate DvD without modifying prompts. The detection method presented in this section can be used as a diagnostic tool for developing such mitigation methods.

\begingroup
\makeatletter
\def\@suffix{S}%
\makeatother
\setcounter{enumiv}{0}
\bibliographystyleS{splncs04}
\bibliographyS{supp}
\endgroup

\end{document}